# Navigating the Edge with the State-of-the-Art Insights into Corner Case Identification and Generation for Enhanced Autonomous Vehicle Safety

Gabriel Kenji Godoy Shimanuki , Alexandre Moreira Nascimento , Lucio Flavio Vismari , João Batista Camargo, Jr. , Jorge Rady de Almeida, Jr. , Paulo Sergio Cugnasca

*Abstract*—In recent years, there has been significant development of autonomous vehicle (AV) technologies. However, despite the notable achievements of some industry players, a strong and appealing body of evidence that demonstrate AVs are actually safe is lacky, which could foster public distrust in this technology and further compromise the entire development of this industry, as well as related social impacts. To improve the safety of AVs, several techniques are proposed that use synthetic data in virtual simulation. In particular, the highest risk data, known as corner cases (CCs), are the most valuable for developing and testing AV controls, as they can expose and improve the weaknesses of these autonomous systems. In this context, the present paper presents a systematic literature review aiming to comprehensively analyze methodologies for CC identification and generation, also pointing out current gaps and further implications of synthetic data for AV safety and reliability. Based on a selection criteria, 110 studies were picked from an initial sample of 1673 papers. These selected paper were mapped into multiple categories to answer eight inter-linked research questions. It concludes with the recommendation of a more integrated approach focused on safe development among all stakeholders, with active collaboration between industry, academia and regulatory bodies.

## I. INTRODUCTION

The automation of vehicle driving is a promising engineering domain. The Autonomous Vehicles' (AVs) intelligent control algorithms have been made possible by the advancements of Artificial Intelligence (AI) research, particularly in Machine Learning (ML) field [1], [2]. It is expected that the wide AV adoption could bring positive societal outcomes such as improving traffic efficiency [3], [4] and reducing accident risks [5], [6].

Waymo represents the state-of-the-art (SOTA) in AV technology and is the only company operating under the toughest permit conditions [7], [8]— 24/7 service, inclement weather, and speeds up to 65 mph—while maintaining a strong safety record [9]–[11]. According to the National Highway Traffic Safety Administration (NHTSA), Waymo's AVs were involved in 415 reported incidents between June 2021 and June 2024 [12], [13]. However, in a detailed analysis covering over 22 million miles driven through June 2024, Waymo reported an 84% drop in crashes involving airbag deployment, a 73% decrease in injury-causing crashes, and a 48% reduction in police-reported crashes compared to human drivers in similar environments [14]. Notably, many of these incidents involved human drivers colliding with Waymo vehicles, such as rear-end collisions or running red lights [14].

Although Waymo has achieved impressive results, that is not the overall current stage of development of the AV industry. Waymo's operation is a single small-scale operation under a pilot program in a few cities [15], and its driverless taxis currently operate in San Francisco, Phoenix, and Los Angeles. However, several companies that once pursued AV technology have resisted or abandoned their efforts due to the immense technical challenges and high costs involved. For example, GM Cruise recently halted its operations due to safety concerns [16], Apple discontinued its electric autonomous car project after years of investment [17], Argo AI (Ford and Volkswagen's joint venture) was shut down [18], and Uber sold its AV division to focus on profitability [19]. In China, Didi Chuxing ceased its AV testing in California amid regulatory challenges [20], TuSimple wound down its U.S. operations before suspending its activities in China due to financial constraints [21], and Qiantu Motors declared bankruptcy despite its new energy vehicle license [22]. Meanwhile, Tesla cars in Autopilot mode have been involved in 956 crashes between 2018 and 2023 [23], [24], 29 of those accidents resulted in fatal crashes [23]. In China, Tesla also faced public backlash over safety concerns, culminating in lawsuits against customers who publicly criticized its Autopilot system [25]. Despite improvements, public reluctance remains [26], [27], and continued accidents could erode trust in AVs, delaying large-scale adoption [28], [29], reinforcing that AVs are not ready for large-scale deployment.

AVs are safety critical systems. Their failures can result in severe consequences, including loss of life and financial or environmental damage [30]–[32]. Therefore, their systems' safety and reliability must be rigorously ensured. However, ensuring AI-based systems (like AVs) reliability and safety performance requires extensive development datasets that contain all the complexities of real-world situations in which AVs will operate, including possible traffic scenarios and vehicle configurations. Real-world data collection and testing [12], [33]–[35] are risky, costly, time-intensive, and insufficient to cover rare but critical scenarios [30], [36]–[38]. This gap between training data and operational conditions increases the risk of unpredictable AV behavior, particularly in novel or hazardous situations. Furthermore, the black-box nature of Deep Learning (DL) models - the most used ML model in AV development [1], [30], [39] - complicates validation, making it difficult to guarantee the safety of decision-making AV driving tasks [39]–[41]. As a result, AVs may struggle with unexpected events – known as 'Corner Cases' (CC) – limiting their trustworthiness and delaying widespread adoption [37], [42]–[44]. To address these challenges, researchers are advancing CC data synthesis, along with simulation-based approaches, to improve model robustness and enhance AV safety.

Indeed, synthetic scenario generation and exhaustive testing in simulated environments appear to be among the key strategies enabling AV advancements, as demonstrated by companies like Porsche [45], NVIDIA [46], and Waymo [47]–[50]. However, while industry leaders achieve remarkable safety improvements, much of their methodology remains undisclosed to maintain competitive advantage, which can foster public distrust in the technology's transparency and trustworthy [51], [52]. As a result, although research on CC is actively pursued in academia, progress is fragmented across various institutions, lacking a unified framework or standardized benchmarks [53]–[55]. This gap limits the ability of researchers and industry practitioners to collaboratively enhance AV safety at scale [53]. Given that AVs are safety-critical systems, advancing an open, structured, and rigorous approach to CC generation is not only desirable but essential [53].

To address this challenge, this study presents a comprehensive Systematic Literature Review (SLR) consolidating the SOTA on CC generation for AVs. To the best of our knowledge, no existing study provides a comprehensive review of CC generation methodologies while also addressing the broader implications of synthetic data for AV safety and trustworthy. This study examines the strategies used for CC, identifies current gaps and limitations, and outlines a research agenda that fosters collaboration between academia and industry. By doing so, it provides a structured overview of the field, supporting a more informed and transparent approach to AV safety validation.

This study is structures into six sections. Section II presents the related work. Then, section III presents the methodology. Section IV presents the results. Finally, Section V provides concluding remarks on the findings.

## II. RELATED WORK

Several SLRs have examined safety, testing, and critical situations in the development of AVs, offering broad perspectives on these topics [56]–[62]. However, despite the growing interest in CC generation, these do not explicitly focus on the role of synthetic CC data, its generation methods, or its limitations and challenges.

Some studies provide broad analyses of safety and testing. For instance, [56] reviews software verification and validation (V&V) for AVs, examining safety standards and regulations. It evaluates the V-model from ISO 26262, highlighting its limitations for ML models, particularly the challenge of avoiding overfitting and the need for continuous retraining with new CCs. Similarly, [57] reviews DL approaches in end-to-end autonomous driving, detailing sensor inputs, system outputs, and safety enhancement techniques such as search-based testing and Generative Adversarial Network (GAN) based attacks. While these works provide valuable insights, they do not emphasize CC generation.

Other studies focus on risk assessment metrics that evaluate safety criteria for AVs [58], [59]. While these metrics are essential tools for assessing safety, they are not explicitly discussed in the context of CC generation, leaving a gap in understanding how such metrics can contribute to scenario creation.

Some works specifically address CC generation. [60] categorizes testing methods in high-fidelity simulators based on SAE levels of autonomy, offering a broad analysis of scenario search techniques. However, its lack of a systematic review methodology limits its reproducibility and comprehensiveness. [61] presents a taxonomy for critical scenario generation algorithms, categorizing them into data-driven, adversarial, and knowledge-based methods. However, the study lacks the rigor of a SLR, missing clear inclusion criteria, reproducible search strategies, and quantitative analysis of trends. It also fails to systematically evaluate method effectiveness or provide structured performance comparisons, limiting its utility as a comprehensive literature synthesis. [62] conducts a systematic mapping study on critical scenario identification, introducing a taxonomy that categorizes methods based on trajectory search and reasoning methods. While their classification provides a structured overview, the study is limited by its ocus on identifying existing critical scenarios rather than generating new corner cases essential for AV validation. Additionally, it lacks a systematic evaluation of the effectiveness of these methods, making it less useful for guiding practical advancements. Although their review spans 2017–2020, it overlooks more recent developments in scenario-based validation and corner case synthesis, limiting its relevance to the rapidly evolving field.

These gaps in the literature highlight the need for a comprehensive SLR on CC generation, particularly focusing on CC data synthesis. This study systematically examines the methods for CC generation, their applications, and the challenges involved, complementing prior work by extending the analysis beyond scenario identification to methods for generating synthetic data used in testing and validation.

## III. RESEARCH METHODOLOGY

The SLR was conducted to comprehensively and systematically identify, collect, filter, and analyze the SOTA techniques for CC identification and generation, highlighting its achievements and limitations, while also understanding the role of CC data, and providing a foundation for further research. The methodology, adapted from [30], [63], seeks to obtain an unbiased synthesis of the literature, as detailed in the structured protocol illustrated in Figure 1.

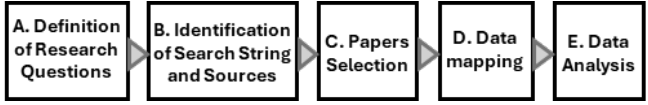

Fig. 1: Literature review's protocol

### A. Definition of Research Questions

The first step was to define the research questions (RQs). Since the study provides a comprehensive analysis of the SOTA CC data synthesis, examining current perspective, future directions, and their implications, the following RQ were proposed:

- **RQ1:** What is the main objective of the studies?
- **RQ2:** What methods are used to identify or generate CCs?
- **RQ3:** What role do CCs play in autonomous system safety?
- **RQ4:** What are the key findings and contributions of the studies?
- **RQ5:** What simulators or datasets are used?
- **RQ6:** What research gaps are identified?
- **RQ7:** What limitations are discussed?
- **RQ8:** What future research is recommended?

### B. Identification of Search String and Sources

The search strategy was designed to identify relevant studies using specific search terms to query multiple indexing databases. The databases utilizes included IEEE explore, Association for Computing Machinery (ACM), Engineering Village (EV), Scopus, SpringerLink, and Web of Science (WOS). No publication date restrictions were applied.

The search string was developed around three core concepts: **High-Risk Scenarios**, **Generation and Simulation Tools**, and **AV Systems**. Each main concept was expanded with relevant related terms to maximize coverage. For instance, synonyms for **High-Risk Scenarios** included "*corner case*", "*dangerous scenario*", "*traffic violation*", "*safety violation*", and "*edge case*". Similarly, **AV Systems** included terms such as "*autonomous drive*", "*self-driving*", "*autonomous vehicle*", "*ego vehicle*", and "*driverless vehicle*", while **Generation and Simulation Tools** encompassed terms like "*generator*", "*framework*", "*search*", "*generation*", "*creation*", and "*simulation*". Figure 2 presents the Boolean structure that organizes these terms into a systematic search query.

To refine the search string, multiple tests were conducted across different databases, using Google Scholar as a validation tool. However, since Google Scholar retrieved only duplicate papers already covered by at least one of the six databases without adding new articles, it was excluded from the final data sources. The increasing volume of studies in this field, as shown in Figure 3, further reinforces the need for a comprehensive review that synthesizes recent advances.

### C. Papers Selection

The initial search returned 1,673 studies across all databases, as summarized in Table I. To refine this broad set of results, a multi-step filtering process was implemented. Initially, the first step was to guarantee that only the studies with the Title-Abstract-Keyword (TAK) fields returning positive to the Boolean search string would be selected. In this process, the metadata of each paper was collected and listed in a Excel spreadsheet. To effectively select the appropriate paper, an evaluation of TAK fields was developed in a *python* script. Through this check, the number of papers was reduced to 430.

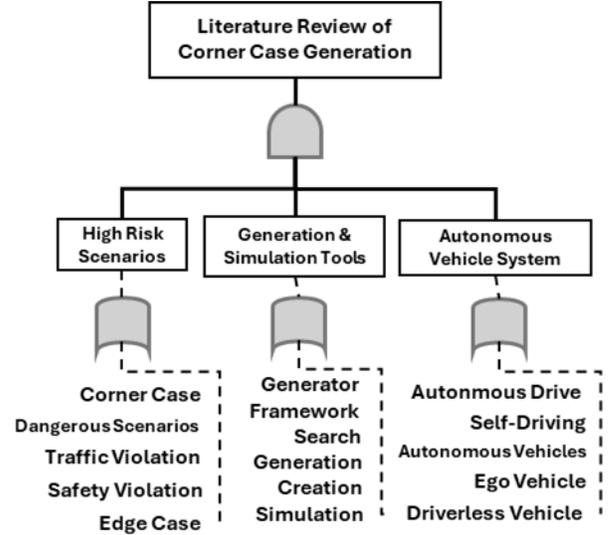

Fig. 2: Summary of the logical structure of the Literature Review search

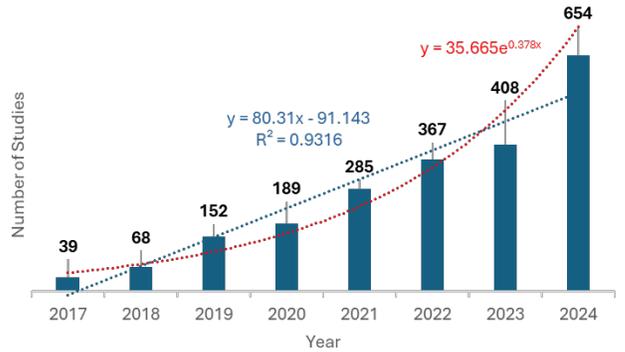

Fig. 3: Trend of retrieved studies per year from Google Scholar

**TABLE I:** Paper count at each filtering stage in the Literature Review protocol

| Step | Scopus | IEEE | ACM DL | EV | Springer | WOS | Total |
|---|---|---|---|---|---|---|---|
| Raw | 200 | 46 | 289 | 81 | 1023 | 34 | 1673 |
| Abstract Reading | 126 | 46 | 213 | 26 | 15 | 4 | 430 |
| Full Paper Reading | 55 | 27 | 47 | 11 | 9 | 1 | 150 |
| **Final Scope** | **40** | **21** | **37** | **4** | **7** | **1** | **110** |
| Database Effectiveness | 20% | 46% | 13% | 5% | 0.7% | 3% | 6.6% |

Then, those 430 filtered studies were further examined with abstract reading, where 280 studies that were not related to the research objectives were excluded. The remaining 150 papers were fully reviewed leading to 41 additional studies exclusions for various reasons. Of these, 18 studies were inaccessible to the authors, one doctoral thesis and two SLRs were excluded because they were not peer-reviewed articles developing a method on the topic; and 20 studies were excluded because they were short articles, research proposals, non-peer reviewed or promotional materials. The final set consisted of 110 relevant studies.

As observable in Table I it is worth noting that IEEE was the most effective database, in which 48% of the retrieved articles were, in fact, incorporated into the analysis of this SLR. In other words, one in every two IEEE articles was relevant to the present study. In comparison, IEEE was 2.4 times more precise than the second best data source (Scopus), and almost 60 times better than Springer. Thus, IEEE venues seem to be the best places to publish and search about the SOTA of AV development.

### D. Data Mapping

A data mapping procedure was executed to systematically extract, map and encode the information extracted from the papers to answer each RQ. At this stage, the first step involved reviewing these 110 papers and writing the answers to each RQ in plain text within a spreadsheet. Next, based on each group of RQ annotations, the authors identified relevant patterns and trends that drove the development of initial encoding categories. The abstraction level was adjusted until it was able to encompass and encode all of the extracted information into a few meaningful categories, at which point a consensus among the authors was reached. The categories are shown in Table II with their correspondence to each RQ. Then, we classified each paper using the established criteria. To ensure that the RQ's answers were consistently discussed and evaluated with homogeneity and fairness, each RQ was classified separately. Uncertainties and disagreements were flagged for further discussion until a final agreement was reached.

The entire process was documented in an Excel spreadsheet, with each paper categorized, annotated with important remarks, and tagged for review as needed. This procedure resulted in a structured dataset containing all classified studies that were aligned with the encoding schema and could fully support the answers of the RQs.

### E. Data Analysis

After the encoding, both quantitative and qualitative analyses were performed to extract insights from the chosen studies. The objective of the quantitative analysis was to provide a summary of the distribution of study topics, methodologies, and important findings. By computing the percentage of papers allocated to each classification, it was possible to find dominant trends, gaps in the literature and underrepresented areas. In addition, a qualitative analysis was performed to synthesize insights the literature associated with each classification, seeking to obtain a broad understanding of the different study types.

These analysis generated the necessary information to answer each RQ. The results of this process are systematically reported in the following section, where each RQ is addressed based on the findings.

Complementarily, the qualitative analysis was performed to interpret and consolidate the key ideas within each category. We synthesized the contributions, methodological methods, and outcomes of studies that were assigned to particular classes by looking at their content and placing them within the broader research landscape. Through this procedure, we were able to comprehend the ways in which various study types help to address each research RQ.

## IV. RESULTS

This section presents the findings and insights resulting from analyzing the selected studies. Each subsection the findings answering the corresponding RQ.

### A. Purpose of the Studies (RQ1)

In general, the studies' primary goals were identifying, generating, or testing AV systems in CC scenarios. Those topics seek to enhance the AV safety, a key challenge in autonomous driving research. While human drivers rely on prediction and reflexes to mitigate accidents [64], handling critical situations effectively remains one of the most complex difficulties in deploying AVs. Consequently, the literature emphasizes developing methods that improve the robustness of control systems through comprehensive and systematic testing which is quite challenging when incorporating ML and DNN [65]–[67]. To achieve this, one of the solutions utilized by the research community focuses on systematically identifying and generating CC scenarios where AV systems are likely to fail [68]. Commonly, the identification of these CC scenarios is a prerequisite for generating test data that evaluates AV systems under diverse and challenging conditions [69]. Among the 110 selected papers, 87 studies (80%) identify or generate CC. Of these, 25 (23%) focus solely on identification, 30 (28%) on generation, and 32 (29%) on a combination of both tasks (integrated approach). Going further, in 80 of these studies (73%) testing the AV safety is listed as a goal. Of these, 15 (14%) uses an identification approach, 23 (21%) uses generation, and 21 (19%) employ an integrated approach. Thus, in most studies, regardless of the approach adopted, the problem to be solved is related to the testing of AVs systems.

*1) CC Identification:* Studies on identification propose methodologies to discover CC that AV systems cannot handle properly. Identifying these CC is essential to prevent potential defects that could lead to severe accidents [69]–[72], as ensuring the safety of AV systems are still an open challenge [73]–[78]. Usually those studies rely on the combination of search algorithms and simulation-based environments [75], [78]–[82]. In fact, the role of virtual simulation environments in improving test coverage and addressing complex conditions is also emphasized in the analysis of these methods [83]–[85]. Thus, these findings highlight the central role of simulation-based in discovering CC to enhance AV safety.

TABLE II: Encoding Categories and Their Correspondence to Research Questions

| RQ | Categories |
|---|---|
| 1 | CC Identification; CC Generation; Integrated Approach & AV Testing |
| 2 | Optimization & Search; Machine Learning; Formal Methods & Reasoning; Testing & Attacks; Algorithms & Miscellaneous Techniques |
| 3 | Testing & Validation; Training & Optimization; Dataset Generation & Augmentation; Safety Assessment & Standardization; Bug Detection & Debugging |
| 4 | Data Generation, New Frameworks, Requirement-Driven & Safety Specification, Real-World Data Integration & Accident Reconstruction, Dataset & Data-Generation, Performance, Benchmark & Comparative Study; Formal Verification and Model-Based Testing; Software, Tools & Platforms |
| 5 | Simulators and Datasets |
| 6 | Scenario & CC Generation; ML-Specific Testing & Data Generation; Coverage & Scalability Methods; Search-Based & Optimization Approaches; Adversarial & Stress Testing; Simulation Realism & Sim2Real Gap; End2End System Validation and Tooling; Formal Methods & Ontologies |
| 7 | Simulation Issues ⇒ Reproducibility; Performance; Domain Gap<br>Scenario and Data Limitations ⇒ Scenario Realism & Complexity; Data Limitations<br>Methodological and Analytical Constraints ⇒ Evaluation Metrics; Parameter Tuning & Combinatorial Issues<br>Generalizability and Validation ⇒ Generalizability; Scenario Validation<br>Safety and Behavior ⇒ NPC Behavior; Safety and Collision Handling |
| 8 | Optimization, Refinement & Methodological Enhancements; Expansion to New Domains & Cross-Domain Applications; Integration of Technologies & Framework/Tool Development; Scenario Complexity, Realism & Data Generation/Management; Addressing Limitations & Biases; Validation & Real-World Testing |

*2) CC Generation:* Studies on CC generating seeks to create realistic yet challenging scenarios that expose AV limitations. Generating such cases is crucial for improving AV safety, as real-world driving involves rare, high-risk events that conventional testing may ignore [86]–[90]. One of the main challenges is balancing realism and diversity while ensuring statistical validity in simulations. To address this, some methods are proposed, such as adversarial learning to create CC [86]–[90], modeling real-world event distributions to improve simulation fidelity and test coverage [89], and structured scenario synthesis methods to generate diverse yet representative driving interactions [91]–[98]. Further development on methods to generate CC are detailed in subsection IV-B (RQ2).

*3) Integrated Approach and AV Testing:* Other papers seek solutions for testing and validating based on identification and generation of critical scenarios. Some of them identify and extract critical scenarios that can expose potential failure points in AV systems [99], [100]. In addition, three studies propose the usage of formal reasoning and computer vision techniques to refine and identify unsafe scenarios from existing safe maneuvers in real world traffic accidents [68], [101], [102]. Another significant subgroup of studies in this category seeks the development of automated and scalable methods to generate complex test scenarios [37], [103], [104]. Rather than relying on manual scenario design, these studies propose automated approaches that leverage adaptive stress testing, formal reasoning, or scenario-specific test suites to generate a wide range of critical scenarios [36], [105]–[109].

Some papers seek to address the broader challenges of achieving comprehensive testing coverage for AV systems, which include simulating realistic and diverse environments, managing the vast parameter spaces involved in scenario generation, and working toward generating scenarios that better align with real-world conditions [110]–[113]. To tackle these challenges, several studies propose new frameworks and methodologies that incorporate road structures, dynamic objects, and multi-agent interactions into scenario generation [114]–[118]. Additionally, some emphasize stochastic and simulation-based testing frameworks capable of handling the inherent uncertainties and non-deterministic nature of real-world driving environments, aiming to identify potential failure patterns and assess AV system robustness in scenarios beyond standard testing [38], [119]–[123]. Further, alternative testing techniques have been explored, particularly for safety-critical components such as perception and machine control modules. These approaches leverage methodologies like rare-event sampling, combinatorial testing, and semantic mutation testing to support AV systems in handling CC and unexpected situations [104], [109], [112], [116], [124]. Collectively, these studies contribute to ongoing efforts in AV testing by exploring different strategies for identifying, generating, and evaluating scenarios that can enhance system reliability in diverse conditions.

### B. Strategies for CC Search and Generation (RQ2)

Five primary approaches of CC generation and identification were found. The list is composed of optimization and search methods (36 studies, 33%), ML-based approaches (54 studies, 49%), formal methods (17 studies, 15%), testing and attack approaches (28 studies, 25%) and miscellaneous techniques (8 studies, 7%). In this analysis each paper may fall into multiple categories, thus these categories are not mutually exclusive, leading to a cumulative percentage exceeding 100%. In general, optimization and search methods focus on systematically exploring parameter spaces to find rare and challenging cases, ML techniques create or detect CCs, formal methods seeks to enhance robustness, testing and attack strategies expose vulnerabilities, and miscellaneous techniques target overlooked CCs. The following sections provide a detailed analysis of these categories, their methodologies, and their applications. Table III summarizes the key techniques associated with each category.

*1) Optimization & Search (33%):* The 36 papers classified in this category focus on optimization and search techniques, both heuristic and numerical. A prominent class of algorithms is evolutionary search methods. Firstly, the most discussed technique in the literature is genetic algorithms (GAs). GAs are employed to evolve and optimize test scenarios by selecting, crossing, and mutating existing cases to generate new, more critical ones. Numerous studies highlight the effectiveness of GAs in creating CCs [71], [79], [80], [103], [108], [112], [123], [128], [130], [131], [133], [153], [163]. These algorithms are useful for identifying rare and extreme scenarios, such as uncommon collision situations or unexpected vehicle interactions, which traditional methods, such as random search, might miss. GAs are often paired with multi-objective optimization strategies [83], [128], [130], [133], which balance trade-offs between parameters like safety, efficiency, and complexity, enabling the generation of test cases that assess both overall system performance and behavior in critical conditions. In addition to GAs, other evolutionary computation methods are also utilized. For instance, novelty search and many-objective optimization are applied to explore the solution space for test scenarios [66], [80]. Unlike traditional search algorithms, novelty search prioritizes diversity over a single fitness goal, uncovering previously unrecognized CCs. Techniques like MAP-Elites [66] further explore this by identifying the most anomalous or misbehaving solutions, aiming to find a more comprehensive testing of autonomous systems.

TABLE III: Summary of Strategies for CC Data Synthesis

| Category | Techniques | References |
|---|---|---|
| **Optimization and Search** (33%) | Genetic Algorithms (GA), Multi-objective optimization, Novelty search, Simulated annealing, Particle swarm optimization, Scenario parameterization, Dimensionality reduction, Search-based testing, Combinatorial testing. | [66], [71], [79], [80], [83], [85], [103], [108], [108], [112], [118], [120], [123], [125]–[134] |
| **Machine Learning** (49%) | Reinforcement Learning (RL), Adversarial learning, Deep RL, Generative Adversarial Networks (GANs), Natural Language Processing (NLP), Monte Carlo simulations, Clustering (k-means, DBSCAN), Gaussian Mixture Models (GMM), Recurrent Neural Networks (RNNs), Adaptive stress testing. | [37], [42], [65], [68]–[70], [72], [74], [76], [77], [87]–[90], [92], [95], [97], [100], [102], [104], [107], [109], [110], [113], [135]–[146] |
| **Formal Methods** (15%) | Model checking, Theorem proving, Logical reasoning, Temporal logic, Hybrid automata, Safety verification, Reachability analysis. | [36], [91], [94], [96], [101], [115], [147]–[152] |
| **Testing and Attack** (25%) | Fuzz testing, Fault injection, Adversarial attacks, Black-box testing, White-box testing, Scenario-based testing, Coverage-guided testing. | [38], [67], [73], [75], [81], [84], [86], [93], [98], [105], [106], [111], [114], [116], [119], [122], [124], [153]–[163] |
| **Miscellaneous** (7%) | Expert-driven scenario design, Data augmentation, Edge-case mining, Traffic simulation-based scenario generation, Stochastic modeling. | [78], [82], [99], [121], [154], [164]–[166] |

Other optimization algorithms, such as simulated annealing [71] and particle swarm optimization [85], are also employed to refine the search space iteratively. These methods guide the search toward regions more likely to reveal vulnerabilities, complementing search-based approaches and enhancing the robustness of the testing process. Complementary techniques like Scenario parameterization and dimensionality reduction techniques, such as parameter sweeping [125], systematically vary key parameters like speed, acceleration, and relative position to identify failure conditions. These methods, when combined with surrogate models [167], which approximate and optimize complex system performance, are particularly effective for generating critical scenarios under dynamic, real-world driving conditions. Some studies also explore the integration of ML techniques, such as reinforcement learning (RL) and fuzzing, with optimization methods to generate edge-case scenarios [122]. These approaches are discussed in more detail in the ML subsection IV-B2.

Finally, search-based software testing and combinatorial testing techniques are employed to systematically explore the input space of ADSs [93], [112], [114], [134], [159]. These methods aim to maximize test coverage, increasing the likelihood of identifying CCs that could lead to system failures. By optimizing the selection and combination of test parameters, combinatorial testing techniques help diagnose CC with a higher probability of triggering faults.

*2) Machine Learning (49%):* The 54 papers categorized under ML for identifying and generating CCs in ADSs present a diverse range of techniques. These studies leverage methodologies such as RL, GANs, DL, Natural Language Processing (NLP), and hybrid approaches to simulate and identify CCs.

A prevalent theme is the use of RL, often combined with adversarial or generative methods [86]–[88], [90], [98], [104], [107], [109], [110], [137], [143], [155], [168]. The creation of hostile driving environment is an example of RL integration with adversarial technique [86], [88], [143], [168]. Further, deep RL is also used to generate CC, enabling agents to learn from simulated environments and develop policies for high-risk, low-probability events [88], [104].

GANs are another relevant technique used for data augmentation, scenario generation, and edge-case creation [37], [86], [89], [139], [142], [162]. This method generate synthetic but realistic driving conditions, such as transferring images between domains (e.g., daytime to nighttime) to simulate environmental changes that could challenge AVs [37]. They are also combined with DL to create richer datasets covering rare or dangerous traffic situations [98], [162], aiming to address limitations of traditional data collection methods. Additionally, studies using GANs revealed that they are primarily applied for dataset augmentation rather than real-time simulation testing.

NLP techniques have been investigated for generating realistic and critical driving scenarios, particularly for testing AV systems under rare or hazardous conditions [136], [138], [140]. These approaches often integrate NLP with image processing to enhance the accuracy of scenario simulations. One approach combines NLP with image processing to extract key elements like road geometry and vehicle movement from accident sketches, which could help in creating more detailed simulations [136]. Another method uses NLP to process unstructured police reports, converting textual data into structured simulation parameters [138]. Additionally, image processing and NLP are integrated to generate simulations from accident sketches, combining visual details (e.g., vehicle trajectories) with contextual information (e.g., accident causes) [140]. While these methods demonstrate potential, their reliance on textual and visual data may limit their ability to capture highly dynamic driving behaviors. Nonetheless, NLP-based techniques offer a promising avenue for generating test cases grounded in real-world data.

In addition to more established ML techniques, various miscellaneous approaches also contribute to the generation and identification of CC. Techniques like Monte Carlo simulations and Markov decision processes model dynamic, random events in traffic environments, facilitating the exploration of parameter spaces to generate diverse CCs [68], [97]. Additionally, statistical learning and data-driven methods, such as k-means clustering, density-based clustering, and Gaussian mixture models, are used to identify and classify CCs from large datasets of driving scenarios [68]–[70], [73], [76], [81], [92], [100], [102], [114], [135]. These clustering methods are often combined with DNNs, long short-term memory networks, and recurrent NN to refine and automate the identification of critical states [68], [69], [74], [76], [102], [135]. By learning complex patterns from historical or synthetic driving data, these models can predict and generate failure scenarios, enhancing the realism and complexity of CCs [76], [89], [145], [169]. Although less explored, the incorporation of driver behavior uncertainty may improve the robustness of these methods, which may enable more accurate testing of autonomous systems under diverse and challenging conditions [117]. Finally, coverage-based approaches, such as situation or scenario coverage frameworks, systematically explore the input space to identify critical scenarios [93], [114]. These methods are often enhanced with other ML models, to automate and optimize test case selection. For example, adaptive stress testing uses RL to dynamically sample test cases, guiding the exploration of critical system behaviors under varying conditions [122], [168].

*3) Formal Methods & Reasoning (15%):* The 17 papers in this category employ formal methods to systematically define, generate, and evaluate CCs. These approaches combine theoretical models with computational tools and simulations to ensure the identification of edge cases that reflect real-world complexity [36], [73], [75], [91], [94], [96], [101], [115], [147]–[152], [156], [169], [170].

Various formal methods are applied to generate and validate CC. For instance, Signal Temporal Logic combined with covering arrays is used to evaluate the behavior of ML components and sensors, focusing on CC to identify potential weaknesses in the system [73]. Similarly, Satisfiability Modulo Theory solvers help generate models that satisfy both structural and numerical constraints, facilitating the validation of paths leading to extreme conditions [147], [150], [151]. Hybrid approaches that combine model-based testing and property-based testing offer flexibility in generating abstract test cases that can adapt to different system specifications [152]. To address uncertainty in dynamic environments, probabilistic methods are paired with formal verification, allowing the generation of test cases that are both formally sound and probabilistically robust [96]. Lastly, learning-based methods, such as trajectory generation, aim to simulate human-like non-player character (NPC) vehicle behaviors in virtual simulations, enhancing the diversity of test cases [91].

Simulation tools are commonly used to extend scenarios into specific configurations, such as varying vehicle speeds, positions, and junction setups. These tools help generate CCs that assess risky driving situations, including potential traffic rule violations, through the integration of formal methods with virtual testing environments [115]. Additionally, comprehensive scenario catalogs are created to capture a wide range of traffic situations, which facilitates the automation of CC generation in controlled simulation environments [170].

*4) Testing & Attacks (25%):* This category includes 28 papers that explore techniques for stress-testing AV systems by exposing them to rare, extreme, or unexpected conditions. These techniques aim to reveal vulnerabilities in sensors, decision-making, or response systems, with the goal of helping AVs operate safely within their operational design domains [38], [67], [73], [75], [81], [84], [86], [93], [98], [105], [106], [111], [114], [116], [119], [122], [124], [153]–[163].

Adversarial testing and perturbations, including black-box adversarial attacks, are used to assess AV path-planning logic by modifying motion trajectories and LiDAR point clouds, identifying weaknesses in the system's behavior under altered conditions [105]. By combining physics-based models with ML, dynamic driving conditions are simulated to provide valuable insights into AV performance across various real-world scenarios [160].

In addition to these approaches, evolutionary algorithms, particularly evolutionary search algorithms, are commonly employed to generate adversarial testing scenarios. These methods create challenging situations that explore the boundaries of system capabilities, uncovering potential vulnerabilities [163]. Mutation testing, a complementary strategy, introduces techniques like adversarial and congestion-based mutations to further examine failure modes in AV systems [84]. Moreover, GAs are applied to perturb driving maneuvers, revealing safety hazards in AV trajectories [153], while the integration of GAs with ML enhances testing efficiency and aids in the identification of critical corner cases (CCs) [122].

Structured testing frameworks can also support the definition and manipulation of test scenarios. For example, the Scenario Description Language [106] allows researchers to inject faults, such as sensor failures or adverse weather conditions, into test scenarios, potentially aiding in the creation of varied and realistic CCs. Similarly, a synthesis-guided falsification framework models system dynamics, including sensor interactions and sensor fusion, which might help generate CCs that are mathematically sound and more closely aligned with the specification of the system [75].

*5) Algorithms / Miscellaneous Techniques (7%):* This category includes 9 papers that employ diverse algorithmic approaches, statistical models, and sensitivity analysis to identify and generate CCs for AV systems. These techniques explore uncommon or extreme scenarios that challenge system robustness, complementing traditional testing methods [78], [82], [121], [154], [164]–[166].

Despite the computational cost one approach uses a multimodal scenario generation based on weighted likelihood maximization and gradient sampling is proposed to create CC [154]. Similarly, randomized search techniques have been employed to set initial conditions for high-speed overtaking sce-narios, offering a promising way to uncover unforeseen CCs [78]. Hardware-in-the-Loop and Vehicle-in-the-Loop systems is proposed to integrate with real-world testing to simulate diverse driving scenarios, relying on the accuracy of traffic data [164]. To enhance testing efficiency, statistical learning methods such as test prioritization are applied. For example, by encoding driving recordings into feature vectors, segmenting them by similarity, and prioritizing these segments based on coverage and rarity, the testing process becomes more focused, although the effectiveness of the method depends on the quality of the encoding and segmentation [165]. Meanwhile, physics-based simulations and sensitivity analysis are also used to examine AV system behavior under specific conditions. Techniques such as finite element simulations and ray tracing investigate radar system vulnerabilities, like terrain-induced interference or ghost targets, although they may not consider all AV sensors [82]. Sensitivity analysis complements this by estimating AV responses to various logical scenarios, helping to determine hazardous conditions before real-world testing, though its accuracy depends heavily on the sensitivity metrics and classification systems employed [121]. Finally, in the domain of vehicle-pedestrian interactions, worst-case models are developed to simulate extreme scenarios at unsignalized crossings, addressing safety concerns [166]. However, these models' reliance on assumptions about negligent behavior may limit their real-world applicability.

## C. CC Application (RQ3)

Five key categories of CC cover the main applications of CCs. These are testing and validation (40 studies, 36%), training and optimization (8 studies, 7%), dataset generation and augmentation (8 studies, 7%), safety assessment and standardization (4 studies, 4%), and bug detection and debugging (10 studies, 9%). A summary of these CC applications can be found in Table IV. Each category is detailed in the following subsections. Notably, in 40 studies (36%), the purpose of CC identification or generation was not explicitly stated, thus they were excluded from this analysis.

*1) Testing and Validation:* CCs are extensively utilized to test and validate ADS, helping to assess their performance, safety, and robustness in rare and critical scenarios. The primary objective is to assess how well an ADS handles rare and critical scenarios, evaluating if it meets safety expectations before deployment. Many studies highlight the use of CCs to validate ADS performance under extreme conditions, benchmarking against regulatory and safety requirements [36], [38], [65], [67], [68], [70], [71], [79], [81], [87], [92], [96], [101], [106], [113], [117]–[119], [121], [125], [127]–[132], [136], [138], [140], [150], [154], [157]–[161], [164], [165], [167], [169].

By simulating these scenarios, researchers can measure the effectiveness of the system, identify performance limitations, and refine models accordingly. Comprehensive testing frameworks based on CCs have been shown to help identify weaknesses and support the effective operation of ADS in diverse environments, potentially accelerating the development process [113], [154], [158]. Additionally, these testing methods contribute to maintaining critical safety standards throughout the system's lifecycle in its operational design domain.

*2) Training and Optimization:* The generation and identification of CCs also serve as a foundation for improving the learning process and optimization of AV systems. Studies focus on using CCs to enhance the training datasets for model control development [86], [105], [107], [142], [144] aiming to improve AV control and safety. For instance, algorithms that fine-tuning with rare scenarios seek to enhance the system's ability to handle unforeseen situations effectively [88], [122]. Training with enriched pedestrian datasets, for example, can directly improve detection and control algorithms, which in turn supports the development of more robust and adaptive ADS policies capable of efficiently responding to critical scenarios [142], [145].

*3) Dataset Generation and Augmentation:* Another significant application of CCs is in generating diverse and comprehensive datasets. Some studies highlight how CCs enrich datasets to capture rare, critical, or unexpected scenarios, which are often underrepresented in standard datasets [74], [102], [135], [151]. This enrichment is essential for benchmarking, developing test cases, and improving ADS algorithms. For example, synthetic data generated from CCs can effectively reduce the cost and time associated with acquiring real-world data, while ensuring extensive coverage of challenging scenarios [91]. Furthermore, the importance of augmenting datasets to improve simulation-based evaluation frameworks is emphasized, as it plays a key role in enabling AV systems to adequately address the variability and complexity inherent in real-world environments [76], [98], [166].

*4) Safety Assessment and Standardization:* CCs are instrumental in assessing the safety of AV systems and defining standards for their performance. [85], [126], [155] illustrate how CCs can be used to evaluate compliance with safety benchmarks and regulatory requirements. For instance, [100] discusses leveraging CCs to supplement existing testing frameworks and establish test schemes that ensure AVs meet stringent safety criteria. This category underscores the critical role of CCs in fostering public trust and regulatory acceptance of autonomous technologies.

*5) Bug Detection and Debugging:* Beyond validation, CCs serve as essential tools for uncovering and diagnosing failures in AV systems. Unlike broad testing, which verifies system behavior, debugging focuses on determining why a failure occurs and addressing its root cause. These studies illustrate how CCs reveal critical flaws in perception, planning, and control algorithms, enabling developers to refine system components before deployment [66], [73], [82], [84], [108], [111], [116], [124], [162], [163]. For example, debugging processes help mitigate failures in advanced driver assistance systems (ADAS) by isolating issues that may not surface under normal testing conditions. [66], [116] Proactively addressing these safety-critical errors enhances system robustness and reduces the risk of real-world operational failures [111], [116]. As emphasized by previous studies, early identification of these vulnerabilities minimizes potential safety hazards, reduces development costs, and streamlines the refinement of ADS

TABLE IV: Summary of CC Applications

| Category | Application | References |
|---|---|---|
| **Testing and Validation** | Corner cases are used to test and validate autonomous driving systems (ADS) in rare and critical scenarios. They help uncover system failures, improve reliability, and ensure compliance with safety standards. | [36], [38], [65], [67], [68], [70], [71], [79], [81], [87], [92], [101], [106], [113], [117]–[119], [121], [125], [127]–[132], [136], [138], [140], [150], [154], [157]–[161], [164], [165], [167], [169] |
| **Training and Optimization** | Corner cases enhance training datasets for reinforcement learning and other models, improving AV control and safety. They fine-tune algorithms to handle unforeseen situations, leading to more robust and adaptive ADS policies. | [86], [88], [105], [107], [122], [142], [144], [145] |
| **Dataset Generation and Augmentation** | Corner cases enrich datasets by capturing rare, critical, or unexpected scenarios. They reduce the cost and time of acquiring real-world data and improve simulation-based evaluation frameworks. | [74], [76], [91], [98], [102], [135], [151], [166] |
| **Safety Assessment and Standardization** | Corner cases assess AV safety and define performance standards. They ensure compliance with safety benchmarks and regulatory requirements, fostering public trust and regulatory acceptance. | [85], [100], [126], [155] |
| **Bug Detection and Debugging** | Corner cases identify and debug vulnerabilities in AV systems. They expose flaws in system performance or design, allowing developers to address issues before deployment and enhance system robustness. | [66], [73], [82], [84], [108], [111], [116], [124], [162], [163] |

technology [111], [116].

### D. Key Findings and Contributions (RQ4)

The primary findings and contributions of selected studies on AV safety and testing can be categorized into eight key areas. Adversarial Data Generation (42%) leads the contributions, focusing on generating extreme scenarios to test AV robustness through techniques like RL and GAs. New Frameworks (41%) introduce innovative simulation, optimization, and scenario-generation approaches to enhance AV safety validation. Requirement-Driven / Safety Specification (17%) emphasizes formalized safety validation through ontologies, scenario prioritization, and compliance testing. Real-World Data Integration / Accident Reconstruction (6%) leverages accident data for realistic test case generation. Performance / Benchmark / Comparative Studies (7%) assess testing methodologies, improving efficiency and reliability. Formal Verification / Model-Based Testing (7%) applies formal methods to validate AV behaviors and critical scenarios. Dataset Construction (5%) seeks to enhance AV perception through synthetic and augmented datasets for rare scenarios. Finally, Software / Tool / Platform (5%) tries to advance simulation environments and automated testing tools. These categories collectively cover the contributions to improving AV safety, trustworthiness, and validation methodologies. Also note that in this analysis each paper may fall into multiple categories, thus these categories are not mutually exclusive, leading to a cumulative percentage exceeding 100%. Finally, Table V summarizes the key techniques associated with each category.

*1) Adversarial Data Generation:* Approximately 42% of the studies in this field focus on Adversarial Data Generation, with the aim of developing adversarial frameworks to generate extreme and challenging scenarios to test the robustness of AV systems [65], [67]–[70], [75], [79], [81], [84]–[86], [88], [90], [92], [95], [97], [102]–[106], [108]–[110], [113], [118], [120], [122], [124], [128]–[131], [133], [137], [139], [143], [153]–[155], [157], [158], [160], [163], [168], [169].

A key contribution is the development of frameworks for adversarial testing, such as ANTI-CARLA, which leverages the CARLA simulator to automate test case creation. These frameworks reveal vulnerabilities in AV systems, even in advanced controllers like Learning By Cheating, which fail under extreme edge cases [106]. Search-based methods, including GA, generate safety-critical scenarios (e.g., pedestrian collisions) more effectively than random testing, uncovering rare failure modes [133]. Adaptive stress testing refines scenarios by targeting critical states (e.g., unsafe behaviors) through data-driven classification, enhancing robustness evaluations [168]. RL methods further advance adversarial testing by guiding systems toward rare high-risk scenarios, reducing reliance on real-world testing [143]. Collectively, these approaches—spanning optimization, GA, and RL—prioritize efficiency and diversity in scenario generation, addressing limitations of traditional predefined or random methods. This enables systematic stress-testing of AVs in realistic adversarial contexts, improving failure identification and safety [69], [95], [118], [122].

*2) New Frameworks:* Studies in this category (41%) focus on the development of new methods and frameworks aimed at improving the safety, testing, and performance of AVs in complex driving scenarios [38], [66], [70], [72], [74], [76]–[78], [80], [82], [86]–[89], [91], [94], [96], [101], [104], [105], [107], [111], [116], [117], [119], [121], [125], [126], [129], [132], [137], [141], [144], [146], [147], [149], [153], [154], [160], [161], [164], [166]–[169], [172]. These contributions span the simulation of critical driving situations and the enhancement of scenario generation methods' accuracy and efficiency.

Key advancements include advanced simulation frameworks integrating tools like CARLA and TensorFlow to detect CC behaviors [141], as well as dynamic scenario libraries combining physics and intelligence for efficient testing [160]. Novel methods address safety-critical events through imitation learning and driver heterogeneity modeling, accelerating AV testing by generating diverse adversarial scenarios [87], [117]. Safety improvements are driven by counterfactual reasoning for perception-failure analysis [169] and semantic mutation testing to uncover LiDAR-based errors [116]. Optimization techniques, such as heuristics applied to naturalistic driving data and formal verification integrated with scenario generation, refine test case relevance and validation efficiency [96], [119]. Practical tools like injection simulation systems and the Sim-on-Wheels framework enhance Software/Vehicle-in-the-

TABLE V: Summary of Key Methods and Contributions

| Category | Key Findings/Contributions | References |
|---|---|---|
| **Adversarial Data Generation** | • Automated adversarial testing frameworks (e.g., ANTI-CARLA).<br>• Search-based techniques (e.g., genetic algorithms) for safety-critical scenarios.<br>• Adaptive stress testing to identify vulnerabilities.<br>• Reinforcement learning for rare-event situations. | [65], [67]–[70], [75], [79], [81], [84]–[86], [88], [90], [92], [95], [97], [102]–[106], [108], [108]–[110], [113], [118], [120], [122], [124], [128]–[131], [133], [137], [139], [143], [153]–[155], [157], [158], [160], [163], [168], [169] |
| **New Frameworks** | • Advanced simulation frameworks (e.g., CARLA + TensorFlow).<br>• Imitation learning (IL) for diverse traffic scenarios.<br>• Counterfactual reasoning for perception failure evaluation.<br>• Heuristics and formal verification for safety-relevant test cases. | [38], [66], [70], [72], [74], [76]–[78], [80], [82], [86]–[89], [91], [94], [96], [101], [104], [105], [107], [111], [116], [117], [119], [121], [125], [126], [129], [132], [137], [141], [144], [146], [147], [149], [153], [154], [160], [161], [164], [166]–[169], [171], [172] |
| **Requirement-Driven/Safety Specification** | • Knowledge-based frameworks for map inaccuracies.<br>• Ontology-based methods for AEB system validation.<br>• Virtual testing for pedestrian collision warning systems.<br>• Tools like AVUnit and RvADS for critical test scenarios. | [73], [76], [79], [83], [93], [111], [112], [114], [115], [121], [123], [131], [149], [150], [156], [159], [165], [166], [170] |
| **Real-World Data Integration/Accident Reconstruction** | • Frameworks like AI2ISO and CSG for critical scenarios.<br>• Tools like CRISCE and AC3R for realistic test scenarios.<br>• Test scenarios for powered two-wheelers (PTWs). | [77], [100], [102], [136], [138], [140], [148], [171] |
| **Dataset Construction** | • Datasets like CODA and Precarious Pedestrian Dataset.<br>• GANs for generating rare traffic scenarios.<br>• Data augmentation for pedestrian detection. | [37], [99], [135], [142], [145], [162] |
| **Performance/ Benchmark/ Comparative Study** | • Genetic algorithms outperform random testing for AEB systems.<br>• SAMOTA method improves testing efficiency for DES.<br>• Comparative studies of CT and SBT for crash scenarios. | [71], [83], [85], [112], [127], [134], [152], [167] |
| **Formal Verification/ Model-Based Testing** | • Automated methods like CriSGen for safety-critical scenarios.<br>• Formal verification for AV perception validation.<br>• Frameworks like A2CoST for plausible scenarios. | [36], [75], [96], [101], [113], [147], [149], [151] |
| **Software/Tool/Platform** | • Adversarial testing frameworks like Sim-ATAV and ANTI-CARLA.<br>• Simulation environments like VISTA 2.0 and AmbieGen.<br>• Tools like RITA for reactive traffic flows. | [42], [73], [98], [106], [122], [163] |

Loop testing by customizing road scenarios and integrating real-world autonomy stacks with simulated elements [94], [161].

*3) Requirement-Driven / Safety Specification:* The studies categorized under Requirement-Driven / Safety Specification, which represent 17% of the total, focus on advancing testing and validation strategies for ADSs and automated vehicles, with a particular emphasis on safety and reliability [73], [76], [79], [83], [93], [111], [112], [114], [115], [121], [123], [131], [149], [150], [156], [159], [165], [166], [170]. These studies address various aspects of testing, from generating test scenarios to refining testing methodologies to ensure compliance with safety specifications required for real-world deployment.

Key contributions include frameworks for generating diverse safety-critical scenarios, such as knowledge-based methods addressing high-definition map deviations [149] and Sim-ATAV's ML-driven CC detection [73]. Formalized specifications and ontologies enhance testing scalability, exemplified by ontology-based validation of Autonomous Emergency Braking systems [156] and black-box prioritization for early fault detection [83]. Virtual testing leverages formal specifications to automate road environment generation, improving efficiency for features like pedestrian collision warnings [150], while safety indicators prioritize scenarios using semantic coverage and rarity metrics [165]. Tools like AVUnit and RvADS advance validation through automated critical scenario generation and temporal logic-based property checking, identifying deficiencies missed by traditional methods [115], [123]. These innovations collectively strengthen safety compliance, enabling rigorous validation of autonomous systems against real-world deployment standards.

*4) Real-World Data Integration / Accident Reconstruction:* The studies in the Real-World Data Integration and Accident Reconstruction category, representing 6% of the total, focus on integrating real-world traffic accident data to enhance safety testing methods for AVs and ADAS [77], [100], [102], [136], [138], [140], [148]. These studies emphasize the generation of realistic test scenarios based on accident data to improve AV safety.

Key contributions include frameworks for Critical Scenario Generation method, which uses computer vision on accident videos [102]. Novel tools such as CRISCE [136], [140] and AC3R [138] enhance simulation fidelity through accident sketches and police report analysis, respectively. These methods generate rare edge-case scenarios [77] and extend to powered two-wheelers using crash data clustering [100], addressing underrepresented conditions and improving testing comprehensiveness for diverse traffic environments.

*5) Dataset Construction:* The studies in this subset, representing 5% of the total, significantly advance AV testing and perception system development by addressing gaps in dataset availability and quality [37], [135], [142], [145], [162]. A key contribution is the development of novel datasets focused on rare or edge cases, which are often underrepresented in conventional datasets. This improves the trustworthiness and robustness of AV perception systems, particularly for object detection and pedestrian recognition tasks. For example, the CODA dataset addresses road CCs, highlighting the limitations of current object detection systems in handling rare objects and facilitating the design of more resilient models [135]. Similarly, the Precarious Pedestrian Dataset covers hazardous pedestrian situations, which are poorly represented in existing datasets, enabling more comprehensive training and testing of AV perception systems [162].

Studies introduce data generation frameworks that employ GANs to synthesize rare scenarios (e.g., sharp cut-ins) to address perception model limitations in long-tail conditions [37]. Data augmentation strategies, such as generating diverse pedestrian poses, enhance detection systems' generalization for underrepresented CCs [142]. These approaches collectively bridge data gaps, improving AV robustness through synthetic and augmented datasets tailored for rare and challenging real-world scenarios.

*6) Performance / Benchmark / Comparative Study:* The studies in the Performance / Benchmark / Comparative Study category, representing 7% of the total, focus on improving the effectiveness, efficiency, and reliability of testing methodologies for automated driving systems (ADS) [71], [83], [85], [112], [127], [134], [152], [167]. These studies primarily address test optimization, safety verification, and system evaluation, directly influencing AV validation processes.

Contributions include a demonstration that GA outperform simulated annealing and random testing in identifying critical scenarios for autonomous emergency braking systems, reducing computational costs [71], and SAMOTA's surrogate models enhancing efficiency in detecting safety violations for DNN-Enabled Systems [167]. Comparative analyses of control systems reveal model predictive controllers surpass polynomial trajectory generators in safety under uncertainty, despite higher computational demands [127]. Test case prioritization methods like MO-SDC-Prioritizer, leveraging multi-objective GAs, excel in large-scale testing environments [83], while Improved Particle Swarm Optimization enhances multimodal critical scenario searches for AV verification [85]. Studies also benchmark combinatorial testing and search-based testing, showing combinatorial testing's strength in combinatorial coverage and search-based testing's efficiency in crash detection, with combined approaches offering robust validation insights [112]. These contributions collectively refine testing efficiency, safety validation, and methodological scalability for ADS.

*7) Formal Verification / Model-Based Testing:* The studies in the Formal Verification and Model-Based Testing category, representing 7% of the total, contribute significantly to ensuring the safety, trustworthiness, and performance of autonomous systems, particularly in cyber-physical systems and AVs [36], [75], [96], [101], [113], [147], [149], [151]. These studies focus on the application of formal methods and automated processes for generating safety-critical scenarios, testing system behaviors, and validating complex system models.

Key contributions include automated scenario generation tools like CriSGen, which creates critical traffic scenarios to stress-test AV behaviors [101], and formal methods for collision/near-miss scenario generation to enhance safety protocols [36]. Integration of formal verification with perception validation is demonstrated via synthesis-guided falsification for rapid safety violation detection in cyber-physical systems [75] and collision-risk estimation in high-fidelity simulators [96]. Novel formal verification frameworks include automated model generation for traffic scenarios using graph-numeric solvers [147], knowledge-based road modeling to improve map reliability [149], and SMT-constrained road network generation for vehicle control software testing [151]. The A2CoST framework further advances critical scenario generation using Answer Set Programming, ensuring plausibility and collision-avoidance [113]. These methods collectively bridge theoretical safety verification with practical deployment, enabling rigorous validation of autonomous systems in simulated and real-world contexts.

*8) Software / Tool / Platform:* The studies in the Software / Tool / Platform category, representing 5% of the total, contribute significantly to advancing testing, verification, and validation methodologies for AVs [42], [73], [98], [106], [122], [163]. These studies focus on the development of simulation-based tools and frameworks aimed at enhancing the robustness, realism, and scalability of AV system evaluations, particularly in complex and dynamic real-world environments.

Adversarial testing frameworks like Sim-ATAV [73], ANTI-CARLA [106], and AutoFuzz [122] automate edge-case generation to provoke system failures (e.g., misclassifications, collisions, traffic violations) under rare conditions. Simulation environments such as VISTA 2.0 [42] enhance sensor-driven policy robustness via novel viewpoint synthesis, while AmbieGen [163] employs evolutionary algorithms to generate adversarial scenarios adaptable across AVs, robots, and drones. Tools like RITA [98] improve traffic flow realism through reactive agent modeling and ML, bridging sim-to-real gaps for safer AV navigation. These contributions collectively address safety, scalability, and real-world applicability challenges in AV testing and deployment.

### E. Simulators and Datasets (RQ5)

Simulation plays an important role in the research of autonomous driving, providing controlled environments for testing and validation. Carla Simulator is the most frequently cited tool, appearing in 21 studies (19%). Its high-fidelity environments, open-source caracteristics, and active community support contribute to its widespread adoption [173], [174]. Following Carla, LGSVL is employed in 12 papers (11%). However, LGSVL was discontinued in 2022 [175], which may affect its future use in research. Other notable simulators include SUMO (5 papers, 5%) [176], an open-source traffic simulation tool; BeamNG.research (4 papers, 4%) [177], a proprietary simulator; and MATLAB/Simulink (5 papers, 5%) [178], a tool commonly used in control design. In general, the selection of simulators in autonomous driving research is influenced by factors such as open-source availability, community support, and simulation fidelity [179], with simulators like Carla and SUMO reflecting a trend toward accessible and collaborative research tools. Other simulators, such as CarMaker, Unity3D, Vista Simulator [42], and highway-env, are also employed but less frequently, reflecting the range of available tools. Table VI summarizes these findings, providing a comprehensive overview of simulation usage trends.

In addition to simulators, a significant number of papers rely on real-world data (17 studies, 15%) or external datasets (10 studies, 9%), with some employing a mixed approach using both simulators and real/external datasets [67], [69], [87], [88], [91], [117], [132], [134], [158], [162]. This indicates the importance of empirical data in the validation of simulation results. Notable datasets include the KITTI [180], COCO [181], and Lyft Prediction datasets [182], which provide valuable real-world scenarios for testing and validation. Additionally, 12 papers (11%) did not specify the simulator used, which may suggest the use of proprietary tools or a focus on theoretical aspects not involving simulation. This diversity in data sources and even the occasional omission of simulator details highlight the varied methodologies and reporting practices in autonomous driving research.

### F. Research Gaps (RQ6)

The analysis categorizes the unresolved research gaps in current CC data generation for AVs, as identified by the studies. The most significant gap is Scenario and CC Generation (53%), where the challenge lies in generating diverse and rare driving conditions. ML-Specific Testing and Data Generation (24%) remains an open issue due to the lack of scalable methods for ensuring robustness against distributional shifts and adversarial inputs. Coverage and Scalability Methods (21%) require better frameworks to enhance scenario diversity and validation. Search-Based and Optimization Approaches (20%) struggle with reproducibility and full scenario exploration. Adversarial and Stress Testing (19%) lacks standardized benchmarks to assess system vulnerabilities under extreme conditions. The Sim-to-Real Gap (12%) remains a significant and practical challenge, with existing simulation techniques failing to replicate real-world complexities accurately. End-to-End System Validation (9%) is underexplored, limiting comprehensive ADS assessments. Finally, Formal Methods and Ontologies (8%) face scalability constraints despite their potential for structured safety validation. Again, observe that in this analysis each paper may fall into multiple categories, thus these categories are not mutually exclusive, leading to a cumulative percentage exceeding 100%. Finally, Table VII summarizes the research gaps associated with each category.

A major gap lies in Scenario and CC Generation (**Category 1**), which, despite accounting for 53% of the research, shows a need for more comprehensive methods for generating diverse and rare driving scenarios [36], [38], [42], [68], [70], [76]–[78], [85], [87], [88], [91], [92], [95]–[98], [100]–[103], [105], [108], [109], [113]–[115], [117]–[120], [122]–[127], [129], [131], [133], [135], [136], [138], [140], [141], [143], [147], [150]–[152], [155], [156], [159], [160], [163], [166], [169], [172]. While these scenarios are essential for evaluating ADS performance under extreme or uncommon conditions, many studies still emphasize the simulation of real-world complexities. These include multi-agent interactions, dynamic environmental changes, and CC that are difficult to replicate through conventional data collection methods [92], [96], [103], [109], [113], [118], [127], [160]. This indicates a need for more effective approaches to better simulate these challenging conditions.

Another notable gap is found in ML-Specific Testing and Data Generation (**Category 7**), which makes up 24% of the studies [37], [65]–[67], [69], [73], [74], [81], [86], [87], [90], [96], [104], [107], [110], [132], [135], [139], [141]–[145], [150], [154], [157], [162], [167]. Research in this area aims to address the challenges of testing ML components within ADS, especially considering the reliance on DL models for perception, planning, and control tasks. These models are prone to issues such as distributional shifts and adversarial inputs, which may affect the trustworthiness of model control, and the lack of diversity in training datasets remains a challenge [139], [142], [144], [145], [162]. While new approaches like neuron coverage metrics [81], [104], [132], domain adaptation [144], [145], [162], and adversarial robustness testing have emerged [73], [110], [141], [154], [157], there is still a lack of scalable solutions that can generalize across varying real-world conditions [96], [107], [143], [144]. This underlines the need for further development in scalable and generalizable testing methods to ensure ADS safety in diverse scenarios.

Additionally, Coverage and Scalability Methods (**Category 2**) are underrepresented, with only 21% of studies addressing this area [38], [68], [76], [77], [79], [81], [83], [89], [92], [93], [100], [112], [115], [121], [125], [126], [146], [147], [150], [159], [160], [164], [165]. Despite the crucial role coverage plays in validating ADS robustness, there remains a significant gap in the development of comprehensive frameworks to measure and improve scenario coverage [115], [125], [126], [147], [150]. Current efforts focus largely on expanding scenario combinations or applying multi-objective optimization [76], [79], [83], [112], [159], but few studies propose innovative metrics or scalable methods that can handle the complexity of real-world driving environments [81], [83], [93], [150], [159]. This gap presents an opportunity for research into flexible, computationally efficient coverage methods that meet the evolving testing needs of both academia and industry.

Search-Based and Optimization Approaches (**Category 6**) represent 20% of the studies but are also an area with potential for further research [71], [76], [79], [80], [85], [101], [103], [106], [108], [111]–[113], [119], [124], [128]–[130], [133], [134], [167], [169]. These methods, which focus on navigating the vast parameter spaces of ADS testing, often rely on algorithms like GAs, evolutionary search, and multi-objective optimization to identify edge cases and enhance coverage. However, their heuristic-driven nature raises concerns about reproducibility and the ability to fully cover the wide range of possible scenarios [71], [101], [112]. This suggests the need for more robust methods that can guarantee the comprehensive exploration of edge cases and improve the efficiency of search-based techniques [85], [128], [133], [167].

TABLE VI: Overview of AV Simulation Tools

| Simulator/ Category | Count | Access Type | Maintenance Status | Official Website | References |
|---|---|---|---|---|---|
| Carla | 21 | Open-source | Actively maintained | carla.org | [36], [68], [75], [81], [84], [86], [93], [96], [102], [106], [109], [111], [113], [122], [141], [143], [146], [154], [155], [167], [172] |
| LGSVL | 12 | Open-source | **Discontinued** | lgsvlsimulator.com | [76], [92], [108], [115], [122]–[124], [128], [131], [133], [153], [165] |
| SUMO | 5 | Open-source | Actively maintained | eclipse.org/sumo | [148] (Foretify), [88], [89], [137], [161] |
| VTD | 4 | Paid | Actively maintained | vires.com/vtd/ | [85], [103], [112], [159] |
| BeamNG.research | 4 | Paid | Actively maintained | beamng.tech/research | [66], [83], [129], [138] |
| CarMaker | 3 | Paid | Actively maintained | ipg-automotive.com | [78], [120], [159] |
| MATLAB/Simulink | 5 | Paid | Actively maintained | mathworks.com/simulink | [97], [102], [114], [120], [169] |
| highway-env | 4 | Open-source | Actively maintained | github.com/eleurent/highway-env | [68], [90], [104], [168] |
| Vista Simulator | 2 | Open-source | Actively maintained | vista.csail.mit.edu | [42], [107] |
| BeamNG.TECH | 2 | Paid | Actively maintained | beamng.tech | [136], [140] |
| Unity3D simulation engine | 2 | Paid | Actively maintained | unity.com | [150], [151] |
| CarSim | 2 | Paid | Actively maintained | carsim.com | [97], [114] |
| Gazebo | 2 | Open-source | Actively maintained | gazebosim.org | [80] (Autorally), [163] (Player/Stage) |
| AVL Model.Connect | 3 | Paid | Actively maintained | www.avl.com | [71], [103], [156] |
| SPAS (Volvo Cars) | 1 | Proprietary | Limited information | N/A | [70] |
| MORAI SIM | 1 | Paid | Actively maintained | morai.ai/morai-sim | [125] |
| OpenDS | 1 | Open-source | Actively maintained | boomaa23.github.io/open-ds | [101] |
| LiDARsim | 1 | Open-source | Actively maintained | github.com/S1ink/UE5-LidarSim | [105] |
| Sim-ATAV (Webots) | 1 | Open-source | Actively maintained | github.com/tuncali/sim-atav | [73] |
| V-REP | 1 | Open-source | **Discontinued** (CoppeliaSim) | coppeliarobotics.com/ | [164] |
| PreScan simulator | 1 | Paid | Actively maintained | tassinternational.com/prescan | [130] |
| PTV Vissim | 1 | Paid | Actively maintained | ptvgroup.com | [120] |
| Unreal Engine 4 | 1 | Open-source/Paid | Actively maintained | unrealengine.com | [169] |
| ANSYS HFSS-FEM | 1 | Paid | Actively maintained | ansys.com | [82] |
| Esmini | 1 | Open-source | Actively maintained | esmini.github.io | [118] |
| LASS (lightweight simulator) | 1 | Open-source | Actively maintained | N/A | [113] |
| SMARTS simulator | 1 | Open-source | Actively maintained | github.com/huawei-noah/SMARTS | [98] |
| NeuralNDE | 1 | Open-source | Limited availability | github.com/zhaoyilin/NeuralNDE | [89] |
| **Not stated** | 12 | | | | [38], [72], [79], [95], [110], [121], [126], [127], [144], [147], [149], [152] |
| Papers using real data | 17 | | | | [105], [135] (UrbanScenarios dataset), [71], [74] (Udacity driving challenge), [157] (Dave, Udacity driving challenge, KITTI), [166] (collected in Sweden), [77], [139], [160], [170] (real car), [37], [94], [100], [116], [119], [142] (real data + external data), [145] |
| Papers using external datasets | 10 | | | | [132], [158] (CIFAR-10, KITTI), [67] (COCO), [87] (Lyft Prediction Dataset), [134] (External database), [69] (Cityscapes dataset), [88], [162] (HighD dataset), [117] (HighD dataset), [91] (InD dataset) |

Adversarial and Stress Testing **(Category 4)** is explored in only 19% of the studies, exposing a gap in research focused on identifying vulnerabilities in ADS through worst-case scenario simulations, such as adversarial attacks targeting perception or decision-making modules [65], [73]–[75], [84], [86], [88], [90], [98], [104], [105], [110], [113], [120], [132], [154], [155], [157], [158], [163], [168]. While these studies often intersect with ML-specific testing **(Category 7)**, adversarial testing uniquely focuses on system weaknesses under extreme conditions. There remains a challenge in establishing standardized adversarial benchmarks [73], [105], [132], [154] and ensuring that these tests accurately reflect real-world risks [75], [86], [90], [155], highlighting the need for further research to develop effective and standardized methods for adversarial testing [74], [104], [110], [163].

In addition, a notable research gap exists in the area of Simulation Realism and the Sim-to-Real Gap **(Category 3)**, which is addressed in only 11.6% of the studies [38], [42], [82], [89], [93], [94], [98], [107], [116], [127], [136], [137], [152]. Despite the advantages of simulators as controlled, cost-effective testing platforms [116], [127], [136], they often fall short in replicating the full range of real-world driving conditions, such as sensor noise, variable weather, and complex agent interactions [107], [116], [136], [137]. While some research has explored advanced techniques like photorealistic rendering, sensor simulation, and domain adaptation to bridge this gap, these efforts are still limited [42], [137]. Many studies continue to rely on standard off-the-shelf simulators rather than pursuing innovative solutions [93], [152]. This underscores the need for further research focused on developing methods that enhance simulation realism and improve the applicability of simulation results to real-world scenarios [42], [94], [116], [127], [137].

A research gap exists in End-to-End System Validation **(Category 8)**, which is explored in only 9% of the studies [83], [84], [106], [111], [118], [122], [128], [148], [153], [161]. These approaches take a comprehensive view of testing by integrating multiple modules and assessing the overall performance of the entire ADS stack. They also focus on developing tools and frameworks for orchestrating complex tests, allowing for the evaluation of interactions between different system components. However, the relatively low focus on this category suggests that many studies continue to prioritize individual components rather than adopting a system-level approach. This underscores a critical need for more research on end-to-end validation, as it is essential to ensure the safety

TABLE VII: Research Gaps Pointed by Authors

| Category | Main Research Gaps | References |
|---|---|---|
| 1. Scenario and CC Generation (53%) | • Need for more comprehensive methods to simulate real-world complexities (e.g., multi-agent interactions, dynamic environmental changes). <br> • Difficulty in replicating rare edge cases through conventional data collection. | [36], [38], [42], [68], [70], [76]–[78], [85], [87], [88], [91], [92], [95]–[98], [100]–[103], [105], [108], [109], [113]–[115], [117]–[120], [122]–[127], [129], [131], [133], [135], [136], [138], [140], [141], [143], [147], [150]–[152], [155], [156], [159], [160], [163], [166], [169], [172] |
| 2. Coverage and Scalability Methods (21%) | • Lack of comprehensive frameworks for scenario coverage. <br> • Limited innovation in metrics and scalable methods for real-world driving environments. <br> • Need for computationally efficient methods. | [38], [68], [76], [77], [79], [81], [83], [89], [92], [93], [100], [112], [115], [121], [125], [126], [146], [147], [150], [159], [160], [164], [165] |
| 3. Simulation Realism and Sim-to-Real Gap (12%) | • Limited ability to replicate real-world conditions (e.g., sensor noise, weather variability, complex agent interactions). <br> • Over-reliance on off-the-shelf simulators. <br> • Need for advanced techniques like photorealistic rendering and domain adaptation. | [38], [42], [82], [89], [93], [94], [98], [107], [116], [127], [136], [137], [152] |
| 4. Adversarial and Stress Testing (19%) | • Lack of standardized adversarial benchmarks. <br> • Difficulty in ensuring tests reflect real-world risks. <br> • Need for effective and standardized methods for adversarial testing. | [65], [73]–[75], [84], [86], [88], [90], [98], [104], [105], [110], [113], [120], [132], [154], [155], [157], [158], [163], [168] |
| 5. Formal Methods and Ontologies (8%) | • Limited application to complex driving environments. <br> • High resource and scalability challenges in developing formal models. <br> • Need for detailed and precise representations of road networks and traffic dynamics. | [72], [96], [101], [123], [147], [149], [156], [170] |
| 6. Search-Based and Optimization Approaches (20%) | • Heuristic-driven methods lack reproducibility and comprehensive coverage. <br> • Need for robust methods to guarantee exploration of edge cases. <br> • Limited scalability in handling real-world complexity. | [71], [76], [79], [80], [85], [101], [103], [106], [108], [111]–[113], [119], [124], [128]–[130], [133], [134], [167], [169] |
| 7. ML-Specific Testing and Data Generation (24%) | • Lack of scalable and generalizable testing methods. <br> • Challenges with distributional shifts, adversarial inputs, and dataset diversity. <br> • Need for innovative approaches like neuron coverage metrics. | [37], [65]–[67], [69], [73], [74], [81], [86], [87], [90], [96], [104], [107], [110], [132], [135], [139], [141]–[145], [150], [154], [157], [162], [167] |
| 8. End-to-End System Validation and Tooling (9%) | • Limited focus on system-level validation. <br> • Need for tools and frameworks to evaluate interactions between system components. <br> • Overemphasis on individual components rather than holistic system performance. | [83], [84], [106], [111], [118], [122], [128], [148], [153], [161] |

of ADS in real-world applications [183], [184].

Finally, Formal Methods and Ontologies (**Category 5**) are explored in only 8% of the studies, highlighting a gap in the research [72], [96], [101], [123], [147], [149], [156], [170]. While these approaches provide a structured framework for specifying traffic rules, scenario structures, and system requirements, making them valuable for ensuring safety and compliance in ADS, their application to complex driving environments remains limited [72], [156], [170]. The high level of detail and precision needed to develop formal models that accurately represent road networks, traffic dynamics, and actor behaviors creates significant resource and scalability challenges [99], [123], [147].

The findings reveal that while significant attention has been given to scenario generation and ML-based testing to address the challenges faced by ADS, critical areas such as end-to-end validation, formal methods, and the sim-to-real gap remain underexplored. This points to the need for more comprehensive research that integrates advancements in scenario generation (**Categories 1 and 6**), robust coverage metrics (**Category 2**), innovative ML testing methods (**Category 7**), and comprehensive validation frameworks (**Category 8**). Addressing these gaps will require a holistic approach, combining interdisciplinary collaboration and emerging technologies to develop scalable and reliable solutions for ADS.

*G. Limitations of Current Work (RQ7)*

The reviewed studies identify key limitations across five major categories: simulation issues (51 studies, 46%), scenario and data limitations (78 studies, 71%), methodological and analytical constraints (42 studies, 38%), generalizability and validation (66 studies, 60%), and safety and behavior (9 studies, 8%). Each category includes distinct challenges that contribute to the overall limitations in AV testing. Also note that each study may fall into multiple categories, meaning these classifications are not mutually exclusive, leading to a cumulative percentage exceeding 100%. Finally, Table VIII summarizes the key limitations associated with each category.

*1) Group 1 - Simulation Issues:* Challenges related to simulation fidelity, computational constraints, and the gap between simulated and real-world environments.

*a) Reproducibility:* A key limitation of simulation platforms is their non-deterministic nature, which affects reproducibility and, consequently, their reliability. The PAIN framework's restricted field-of-view reduces realism in dynamic environments like rear-end collisions [86], while the Apollo platform exhibits outcome variability [66], [108], [133]. Large-scale platforms like Grid'5000 further challenge experiment replication due to accessibility constraints [36]. These factors impede fine-tuning, failure analysis, and system robustness, requiring improvements in simulation determinism and agent behavior realism [36], [108], [133]. Thus, there are opportu-nities to enhance the simulation determinism, improve agent behavior realism.

TABLE VIII: Limitations of Current Approaches in CC Data Synthesis for AVs

| Category | Limitations | References |
|---|---|---|
| **Group 1 - Simulation and Performance Issues** | | |
| Simulation Reproducibility | • Non-deterministic simulations cause inconsistent results.<br>• Reproducibility challenges on large-scale platforms.<br>• Hinders reliability, fine-tuning, and failure analysis. | [36], [66], [86], [108], [133] |
| Performance Constraints | • High computational costs and scalability issues.<br>• Long computation times in complex scenarios.<br>• Limited scenario diversity due to resource-intensive processes. | [72], [74], [75], [93], [96], [101], [105], [106], [108], [110], [131], [141], [148], [151], [153], [163], [167] |
| Domain Gap | • Simulated environments lack real-world complexity (e.g., weather, lighting).<br>• Inability to replicate human behavior.<br>• Synthetic data introduces biases and lacks real-world validation. | [37], [42], [65], [76], [80], [94], [98], [118], [143], [146], [152], [162] |
| **Group 2 - Scenario and Data Limitations** | | |
| Scenario Realism and Complexity | • Limited realism and representativeness in generated scenarios.<br>• Oversimplified test environments neglect urban and adverse conditions.<br>• Fixed parameters reduce diversity and complexity. | [38], [68], [78], [93], [97], [103], [107], [117], [120], [127], [136], [139], [141], [150], [151], [163], [166] |
| Data Limitations | • Insufficient sample sizes and lack of dataset diversity.<br>• Geographic biases and missing labeled corner cases.<br>• Overly specific datasets limit generalizability. | [69], [77], [88], [91], [92], [100], [116], [120], [130], [145] |
| **Group 3 - Methodological and Analytical Constraints** | | |
| Evaluation Metrics | • Traditional metrics (e.g., TTC) fail to capture real-world complexity.<br>• Over-reliance on static metrics neglects dynamic interactions. | [38], [65], [67], [76], [84], [146], [154], [156]–[158], [160] |
| Parameter Tuning and Combinatorial Issues | • Fixed parameters limit diversity and realism.<br>• Model sensitivity to parameter choices introduces uncertainty.<br>• Combinatorial explosion makes comprehensive testing infeasible. | [38], [72], [114], [116], [119], [121], [125], [134] |
| **Group 4 - Generalizability (Genl.) and Validation** | | |
| Genl. | • Limited transferability to real-world conditions.<br>• Overfitting due to narrow or synthetic datasets.<br>• Scalability issues in large-scale environments.<br>• Lack of real-world validation undermines confidence. | [42], [66], [78], [83], [95], [107], [111], [116], [117], [120], [129], [134], [137], [146], [149], [152], [159], [161], [162], [165], [168] |
| Scenario Validation | • Simulated scenarios lack real-world representativeness.<br>• Insufficient real-world validation causes misalignment.<br>• Scalability issues restrict replication of diverse conditions. | [37], [77], [78], [80], [87], [101], [110], [111], [121], [127], [136], [141]–[144], [165], [166] |
| **Group 5 - Safety and Behavior** | | |
| NPC Behavior | • Unrealistic NPC behavior reduces simulation authenticity.<br>• Variability in NPC actions causes inconsistent results.<br>• Constrained action spaces limit scenario diversity and advanced behaviors. | [72], [90], [108], [123] |
| Safety and Collision Handling | • Insufficient handling of safety-critical scenarios.<br>• Constrained action spaces limit collision avoidance diversity.<br>• Reliance on specific datasets restricts generalizability.<br>• Overlooks broader safety aspects (e.g., real-time decisions). | [68], [95], [98] |

*b) Performance:* Many testing frameworks face significant computational burdens, relying on high-performance hardware (e.g., GPUs) and resource-intensive algorithms [72], [101], [106], [153]. Frameworks like AdvSim and AV-FUZZER struggle with scalability and real-time applicability due to computationally demanding processes, particularly for large-scale or high-fidelity simulations [105], [153]. Similarly, methods such as PEPA and AmbieGen face long computation times in complex or high-dimensional spaces [72], [163]. These constraints limit the diversity of tested scenarios, often focusing on simplified traffic models or single-actor perturbations, neglecting complex multi-agent interactions [72], [101], [153]. Furthermore, while these frameworks can identify some CCs, the high computational costs associated with extensive scenario testing make large-scale, practical applications challenging [74], [75], [93], [96], [110], [131], [141], [148], [151], [153], [163], [167]. Thus, there are opportunities to optmize computational efficiency and expand scenario complexity to ensure robust and scalable testing frameworks.

*c) Domain Gap:* A critical limitation is the gap between simulated and real-world environments, which undermines the safety of ADS evaluations. Simulations often fail to capture real-world complexity, such as dynamic weather, lighting, and road conditions [37], [65], [76]. Additionally, the inability to accurately replicate human behavior, such as unpredictable driver actions or pedestrian intentions, further limits the generalizability to find CCs [94], [98], [143], [162]. Tools like CARLA, while valuable, are criticized for their inability to replicate critical edge cases, sensor inaccuracies, or unexpected system failures [42], [118], [152]. Additionally, the reliance on synthetic data (e.g., GAN-generated datasets) introduces biases and fails to reflect real-world complexity [37], [162]. The lack of real-world validation exacerbates these issues, as simulations may not accurately predict system performance in unpredictable settings [80], [94], [143], [146]. Thus, there is opportunity to improve simulation fidelity, with an emphasis to integration of real-world data.

*2) Group 2 - Scenario and Data Limitations:* Challenges related to scenario realism, complexity, and data availability.

*a) Scenario Realism and Complexity:* A recurring limitation across studies is the inability to generate scenarios that are both realistic and representative of real-world conditions, while also ensuring generalizability across diverse environments. For instance, SMT-based road generation methods struggle to formalize non-linear constraints, limiting their ability to model complex intersections accurately [151]. Similarly, models that only employ Time-to-Collision evaluation metric often overlook critical factors such as driver behavior, pedestrian dynamics, and environmental influences, reducing their applicability to real-world traffic scenarios [166]. Frameworks like SitCov and AmbieGen rely on simplified vehicle models and metrics, which fail to capture the diversity and dynamism of real-world driving environments [38], [93], [163].

Furthermore, many studies oversimplify test scenarios by focusing on controlled environments (e.g., highways, tunnels, parking lots) [68], [78], [107], [127], [141], neglecting the complexity and unpredictability of urban settings or adverse weather conditions (e.g., rain, fog, lighting variability) [136], [139], [150]. The reliance on predefined vehicle models [38], fixed parameters [97], [103], [117], or specific accident types [120] further limits the diversity needed to simulate complex interactions, such as those at urban intersections or rare events.

This narrow scope reduces the generalizability of findings and hinders the evaluation of autonomous systems in dynamic and evolving conditions [79], [103].

The development opportunities in this category requires more sophisticated approaches that replicate real-world complexities, incorporate diverse and adaptive behaviors (e.g., NPCs, pedestrians), and ensure robust testing across varied environments. This will enhance the realism, representativeness, and transferability of scenarios, ultimately improving the reliability and scalability of autonomous systems in real-world deployment.

*b) Data Limitations:* The reviewed studies highlight critical challenges in data availability and quality. Insufficient sample sizes often fail to capture rare or extreme events, such as traffic crashes, limiting reliability [69], [77], [91], [100], [145]. Frequently, these datasets lack diversity, omitting key environmental factors like weather or road types [92], [120]. Also, geographic biases further restrict generalizability, as datasets often focus on specific regions (e.g., China, Germany) that may not reflect broader traffic patterns [88], [100]. Another significant issue is the absence of labeled CCs, which are essential for generating robust test scenarios [69], [116]. This gap undermines testing methodologies, as CCs are vital for ensuring AV safety in unpredictable conditions [92], [130]. Overly specific datasets, such as those focused on expressway accidents, also limit generalizability to more complex driving environments [88], [120].

*3) Group 3 - Methodological and Analytical Constraints:* Focuses on evaluation metrics, parameter tuning, and combinatorial challenges.

*a) Evaluation Metrics:* Current metrics often fail to capture the complexity of real-world driving environments. Traditional metrics like Time-to-Collision and headway focus narrowly on basic safety assessments, neglecting factors such as traffic flow, multi-agent dynamics, and environmental variability [38], [146], [154], [156], [160]. The reliance on static or overly simplistic metrics fails to account for uncertainty and real-time interactions within traffic ecosystems [67], [76], [84]. A paradigm shift toward holistic, adaptive, and dynamic metrics is essential to ensure AVs are adequately tested for safety and performance across diverse conditions [65], [157], [158].

*b) Parameter Tuning and Combinatorial Issues:* Many studies rely on fixed or predefined parameter values, limiting the diversity and realism of simulated environments [72], [114], [119], [125]. Parameter optimization is challenging due to the model's sensitivity to parameter choices, which are often based on expert assumptions or heuristic methods, thereby introducing epistemic uncertainty [116], [121]. Manual intervention in simulation tests exacerbates these limitations by introducing bias and reducing scalability [38], [114]. Combinatorial explosion further complicates testing, as the exponential growth of parameter combinations makes comprehensive testing computationally infeasible [121], [134].

*4) Group 4 - Generalizability and Validation:* Challenges related to the transferability of findings and real-world validation.

*a) Generalizability:* A significant challenge is the limited generalizability of findings to real-world conditions. Many studies rely on simulations like CARLA or BeamNG, which fail to capture dynamic real-world elements such as varying weather conditions and complex road geometries [42], [83], [107], [111], [137], [149]. The use of narrow datasets or synthetic data leads to potential overfitting, where methods perform well in specific environments but struggle in new settings [116], [117], [161], [162], [168]. Scalability is another key issue, as methods effective in small-scale environments face computational challenges when scaled to large, heterogeneous traffic scenarios [66], [120], [134], [146], [165]. Real-world validation remains a critical gap, undermining confidence in the transferability of findings [78], [83], [95], [107], [111], [117], [129], [137], [146], [149], [152], [159], [161], [162].

*b) Scenario Validation:* Another major limitation is the lack of real-world representativeness in simulated scenarios. Many studies rely on controlled or simplified environments and synthetic data, failing to capture real-world complexity [78], [101], [127], [136], [141], [144], reducing the efficiency of validation methods [77], [87], [121]. This is compounded by insufficient real-world validation, leading to misalignment between simulation outcomes and actual AV performance [77], [80], [165], [166]. Scalability issues, such as computational limitations in methods like GANs or RL, further restrict the ability to replicate diverse conditions [37], [87], [110], [111], [142], [143].

*5) Group 5 - Safety and Behavior:* Focus on NPC behavior and safety-critical scenario handling.

*a) NPC Behavior:* The unrealistic behavior of NPCs, such as sudden lane changes or disregard for traffic rules, reduces simulation authenticity and undermines the reliability of ADS testing [72], [108]. Variability in NPC actions leads to inconsistent results and false positives, where crashes occur despite no opportunity for the ego-vehicle to react [108]. Constrained action spaces limit scenario diversity, particularly in complex environments like intersections, and hinder the simulation of advanced behaviors such as overtaking [90]. Again, scalability issues also arise in modeling NPC behavior for multi-agent or large-scale environments, reducing real-world applicability [72], [123].

*b) Safety and Collision Handling:* Current methods often fail to adapt to dynamic, real-world conditions, leading to insufficient handling of CC scenarios. For instance, in the RITA environment, policies struggle to simulate human-like responses, resulting in collisions [98]. Constrained action spaces further limit the diversity of CCs, narrowing the scope of collision avoidance [68]. The reliance on specific datasets, such as CRISCE and CommonRoad, restricts generalizability to diverse real-world conditions, particularly in complex scenarios like critical intersections [95]. Broader safety aspects, such as real-time decision-making and traffic flow optimization, are often overlooked [95].

## H. Future Directions (RQ8)

While most of the studies suggest future research directions to address limitations and broaden the scope of AV testing, 16

papers (15%) did not explicitly propose future work, thus they were excluded from this analysis. The remaining studies suggest a wide range of future research categories, which are summarized as follows: optimization and refinement of AV testing methodologies (31 studies, 28%), expansion to new domains and cross-domain applications (25 studies, 23%), integration of advanced technologies and framework/tool development (20 studies, 18%), scenario complexity and realism (18 studies, 16%), addressing limitations and biases (3 studies, 3%), and validation and real-world testing (11 studies, 10%). Table IX summarizes these recommendations for future research in this evolving field.

**TABLE IX:** Future Research Directions in CC Data Synthesis for AVs

| Category | Directions | References |
|---|---|---|
| Optimization, Refinement, and Methodological Enhancements | • Optimize parameters (e.g., GA, SA).<br>• Refine scenarios with real-world elements.<br>• Enhance frameworks for Sim2Real.<br>• Improve efficiency via prioritization and metrics.<br>• Use ML for feature extraction.<br>• Apply rare-event simulations and formal methods. | [38], [67], [71], [74], [80], [83], [85], [87], [95], [96], [102], [103], [106], [111], [113]–[116], [120], [121], [127], [128], [137], [140], [143], [144], [155], [156], [161], [164], [165] |
| Expansion to New Domains and Cross-Domain Applications | • Add diverse traffic participants.<br>• Adapt to urban settings and varied road topologies.<br>• Validate across geographical contexts.<br>• Simulate adverse weather.<br>• Extend to UAVs, AUVs, CPS.<br>• Integrate into industrial frameworks. | [36], [68], [71], [72], [78], [83], [88], [91], [92], [102], [111], [118], [121]–[123], [125]–[127], [129], [144], [145], [150], [151], [155], [170] |
| Integration of Technologies and Framework/Tool Development | • Use metaheuristics for test generation.<br>• Apply ML for automated scenario generation.<br>• Integrate advanced sensors (e.g., LiDAR).<br>• Extend frameworks (e.g., CSG, MOSAT).<br>• Automate test generation.<br>• Combine evaluation and training. | [73], [79], [80], [84], [86], [90], [93], [97], [98], [102], [103], [110], [128], [133], [136], [140], [154], [157], [163], [165] |
| Scenario Complexity, Realism, and Data Generation/Management | • Enhance road generation with real-world data.<br>• Incorporate higher-level metrics (e.g., comfort).<br>• Use GANs and game engines for rendering.<br>• Generate synthetic pedestrian data.<br>• Automate scenario generation.<br>• Integrate video and synthetic datasets. | [73], [86], [88], [90], [91], [100], [110], [112], [117], [118], [122], [129], [133], [145], [148], [160], [163], [166] |
| Addressing Limitations and Biases | • Improve data quality and diversity.<br>• Mitigate biases in predictions.<br>• Address data privacy and ethics.<br>• Handle heterogeneous sensors. | [74], [84], [91] |
| Validation and Real-World Testing | • Conduct field tests (e.g., vehicle-in-the-loop).<br>• Test synthetic data (e.g., fog).<br>• Align metrics with real-world needs.<br>• Optimize surrogate vehicle modulation. | [38], [69], [70], [84], [94], [110], [125], [139], [146], [166], [172] |

*1) Optimization, Refinement, and Methodological Enhancements (28%):* The optimization and refinement of AV testing methodologies to improve efficiency, scalability, and realism are emphasized as key areas for future work in most studies. Important areas include parameter optimization, where heuristic algorithms like GAs, simulated annealing, and particle swarm optimization are proposed to fine-tune parameters and identify critical scenarios more effectively [71], [74], [83], [85], [111], [120], [121], [127], [164]. Scenario refinement is another critical area, with studies recommending the integration of real-world elements such as terrain features, lane configurations, and accident statistics into scenario design to enhance diversity and realism [87], [102], [117], [129], [137]. Framework enhancement is also a priority, with proposals to extend simulation frameworks (e.g., MOSAT [128]) to support additional driving maneuvers, simulators, and dynamic conditions [114], [128], [137]. Research on Sim2Real gap [32] and incorporating mixed-input generation, partially observable states, and sensor noise are also highlighted [67], [116], [143], [155], [161]. Efficiency improvements through test case prioritization, reduction, and metric development are proposed to optimize resource use while ensuring thorough evaluations [38], [83], [121], [156].

Methodological enhancements focus on leveraging ML for automated feature extraction, smoothing algorithms, and dynamic adaptation to operational contexts are also suggested [67], [80], [87], [103], [120], [140], [165]. Employing optimization-based approaches and rare-event simulations to identify failure points in high-dimensional search spaces is also suggested [85], [106], [121]. Extending testing frameworks using formal methods to manage multiple actors and complex road scenarios for terrestrial and non-terrestrial autonomous systems is another suggested direction [95], [96], [113], [115], [120], [144].

*2) Expansion to New Domains and Cross-Domain Applications (23%):* The extension of AV testing frameworks to new environments, scenarios, and problem domains is suggested as a direction for future research in several works. Key directions include the incorporation of additional traffic participants like pedestrians, cyclists, and other vehicles to simulate more realistic traffic interactions [71], [78], [93], [102], [125], [155]. Adapting methodologies to complex road environments such as urban settings, intersections, and varied road topologies is also emphasized to address challenges like multi-lane configurations and high-speed maneuvers [36], [68], [72], [88], [91], [118], [127], [150], [151]. Validating frameworks across different geographical and cultural contexts is recommended to account for variations in traffic patterns, driving behaviors, and regulatory environments [92], [123], [170]. Simulating environmental factors like weather conditions (e.g., rain, fog, snow) and visual obstructions is proposed to improve trustworthiness in adverse conditions [68], [72], [121], [126].

Additionally, extending methodologies to adjacent domains such as connected AVs and intelligent decision-making systems is highlighted [78], [122], [129], [145]. Adapting existing methods like CART [111] and test case prioritization to unmanned aerial vehicles, autonomous underwater vehicles, robotics, and cyber-physical systems is also proposed [83], [144], [167]. Finally, integrating techniques into industrial frameworks like AICAS for broader applicability to safety-critical systems is suggested [83], [100].

*3) Integration of Technologies and Framework/Tool Development (18%):* The integration of advanced technologies into AV testing frameworks is suggested as a focus for future work. Key areas include the use of search algorithms and metaheuristics, such as GAs, simulated annealing, and quality-diversity search techniques, to improve test case generation and scenario diversity [103], [136], [140], [163]. ML is proposed for automated scenario generation, feature extraction, and failure-revealing test case discovery [79], [93], [133], [163], [165]. Integrating advanced sensor technologies, such as multiple cameras and LiDAR, is suggested to enhance perception and data collection [86]. Extending adversarial testing to unconventional surfaces and dynamic environments is also recommended to uncover edge cases [90], [97], [157].

Another prominent direction is related to Framework and tool development which focuses on extending scenario generation frameworks like Critical Scenario Generation and MOSAT to support more diverse scenarios and integrate with simulators like CARLA [102], [128]. In this context, incorporating dynamic obstacles, diverse road configurations, and adversarial agents to enhance testing environments is proposed [86], [93], [97]. Automating test case generation and improving data handling to streamline testing processes is also emphasized [80], [84], [136]. Combining evaluation and training processes to form adversarial training frameworks for improved safety is another key direction [97], [154]. Additionally, integrating tools like RITA into the SMARTS simulator is proposed to standardize testing across platforms [98].

*4) Scenario Complexity, Realism, and Data Generation/Management (16%):* Some authors emphasize enhancing scenario complexity and realism. Key directions include extending road generation algorithms to include varied lane configurations, real-world terrain data, and dynamic traffic flows [88], [122], [129], [166]. Incorporating higher-level metrics like passenger comfort and maneuver complexity, as well as simulating diverse driving styles, is proposed [112], [117], [118], [160]. Using advanced technologies like game engines, ML techniques, such as GANs, to improve scene rendering and generate failure-revealing test cases is also highlighted [73], [110], [133], [163]. Additionally, modeling long-duration, dynamic traffic flows and incorporating adversarial agents are proposed to enhance realism [90], [97].

To address data scarcity, several approaches are suggested to enhance data generation and management, particularly for pedestrian-related scenarios. One study suggests generating more diverse intersection trajectory data to simulate human-like behaviors in critical traffic situations, which can improve trajectory accuracy and cover a wider range of driving scenarios [91]. Additionally, the use of synthetic data is highlighted as a solution to overcome real-world data limitations, particularly for applications like Fallen Person Detection and AV system training [145]. Automating scenario generation from AV accident reports is also recommended to efficiently expand testing databases and improve data management, ensuring better coverage of pedestrian-involved scenarios [148]. Integrating video data and synthetic datasets is emphasized as a key direction to enhance the accuracy and diversity of pedestrian data in critical traffic situations [86], [91], [100].

*5) Addressing Limitations and Biases (3%):* Limitations and biases in current AV testing methodologies are suggested as future work. Key areas include improving data quality and diversity by expanding datasets and mitigating biases in predictions [74], [91]. Addressing data privacy and ethical concerns is also mentioned, particularly in the collection and use of real-world data [74]. Enhancing the configurability of misbehavior oracles and reducing misclassification biases are recommended to improve the accuracy of behavior assessments [84]. Additionally, addressing challenges posed by heterogeneous sensors and varying resolutions is highlighted as a another area for future work [74].

*6) Validation and Real-World Testing (10%):* Some authors highlight the need for real-world validation to bridge the simulation-to-reality gap. Key suggestions include conducting field tests, such as vehicle-in-the-loop simulations and real-world field tests, to validate theoretical models [70], [84], [94], [110], [125], [146], [166], [172]. Testing synthetic data, such as fog data, is proposed to avoid limitations like false alarms and improve realism [139]. Aligning testing metrics with real-world needs and optimizing surrogate vehicle modulation for more challenging tests are also indicated [38], [69].

## V. CONCLUDING REMARKS

This study synthesizes the current state of research on CC identification and generation for AVs, emphasizing the various methodologies used and the challenges that remain.

This study identified three main objectives in the literature in the domain: identifying existing CCs, generating new ones, and combining both approaches. Identification focuses on finding critical scenarios but may miss other important cases. Generation can introduce novel scenarios, but they may not always be practical or relevant. The studies aiming to identify and combine seek to achieve a better balance, offering more variety but often at a higher computational cost.

Regarding the methods used to identify or generate CCs (RQ2), a variety of approaches have been explored in the literature. Search-based optimization techniques, such as evolutionary algorithms, seem to be particularly effective to uncover rare and extreme scenarios, while ML approaches, including GANs and RL, can create challenging yet plausible conditions. However, these methods may face challenges in accurately representing the full spectrum of real-world scenarios, potentially leading to models that do not generalize well to unseen situations. Formal verification methods offer a mathematically rigorous approach, which may enhance the reliability and safety of autonomous systems. However, they struggle with scalability in complex systems like AVs. Adversarial testing, including fuzzing and perturbation-based attacks, effectively reveals vulnerabilities but lacks standardized frameworks for broader deployment. The diversity of these approaches highlights the need for a hybrid framework leveraging their individual strengths and diminishing their individual weaknesses. Therefore, a proper integration of search-based, ML, and formal methods into a hybrid engine to support and integrated framework for CC generation could enhance the approach robustness and generalizability, ultimately improving system reliability and safety. However, no single study was found to combine multiple approaches.

The use of simulation-based methods for generating CCs (RQ3) is widespread in the literature. That is driven by their efficiency in providing controlled environments for large-scale testing [170]. Tools like CARLA, BeamNG, and SUMO are commonly employed due to their repeatability and cost-effectiveness in scenario reproduction [173], [176], [177], [185]. However, while simulators have advanced, challenges remain in ensuring that generated CCs reflect real-world complexities [186]. The gap between simulated and real-world performance is a significant concern [187], as current simulators often fail to capture the full range of environmental variables and human behavioral unpredictability [37], [38], [65], [76], [93], [146], [154], [156], [160], [163]. This raises questions about the applicability of CCs identified in synthetic settings to real-world AV deployment [186], [188]. Therefore, a proper framework should also rely on real-world data to capture and enable ML algorithms to model human-behavior in real-world situations. Moreover, the framework should support data augmentation leveraging in those ML models to generate synthetic variations. However, in order to restrict the novel scenarios into a universe of feasible, plausible, and relevant (i.e. realistic), for the real-world context, the framework should not rely solely into AI connectionist approaches, such as ML models. In fact, it must constrain those models' outputs using AI symbolic approaches and formal methods (rule based). That is, the generated CC should be filtered by a set of rules to ensure they are realistic, therefore reducing the gap between the simulations and real-world applications of those scenarios.

The main findings and contributions of these works (RQ4) highlight the continued dominance of simulation as the primary approach for testing AVs, but emphasize the ongoing challenge of translating these simulations into real-world applicability. Despite significant advancements in simulation fidelity, sensitivity to initial conditions remains a major obstacle, as slight variations can lead to vastly different outcomes, complicating reproducibility and model validation [189]. The insufficient and scarce real-world CC data further exacerbates this issue, with rare and extreme events being underrepresented in current driving datasets [74], [102], [135], [151]. That data is highly valuable and probably exists as the key of success of private companies such as Waymo and Tesla. They can create competitive advantages for the existing players possessing it since they can support the advancement of AV systems towards the higher SAE Levels. For that reason, they used to be not available to the general public. This delays the development of robust models capable of handling such scenarios. Additionally, inherent biases in the collected data, stemming from geographic limitations or sampling imbalances, pose further challenges for model generalization, reducing the effectiveness of learned models when applied to diverse environments. These limitations underscore the need for more advanced testing frameworks that incorporate real-world data and improve simulation accuracy to ensure AVs can safely navigate complex, unpredictable situations.

Several key research gaps (RQ6) were identified in the literature. A significant gap is the lack of standardized definitions and taxonomies for scenario and CC generation, which impedes comparisons and benchmarking across studies and limits the effectiveness of methodologies in addressing critical driving situations. Another gap is the disparity between simulation and real-world conditions, as current simulation models often fail to reflect the complexities of the real world, affecting the applicability of generated CCs. Additionally, the scarcity of high-quality, publicly available datasets focused on CCs exacerbates the problem, as even existing datasets, though focused on rare and extreme events, are still insufficient to capture the full range of such scenarios.

Furthermore, the metrics used to evaluate these scenarios are often insufficient, with simplified proxies like time-to-collision or crash frequency, failing to capture the complexities of multi-agent interactions and other real-world traffic dynamics. This issue is prevalent in fields analyzing rare events. For instance, many studies on ML models for predicting or classifying rare events—such as software defects—rely solely on accuracy to assess model quality [190], [191]. However, these studies often overlook the severe class imbalance in their datasets [190], [191]. For example, when defective software modules comprise only 1% of the data, a model that labels all modules as non-defective can achieve 99% accuracy, misleadingly suggesting high performance. This flaw has been widely reported in published ML studies [191], which often neglect more robust evaluation metrics for rare events, such as precision, recall, the F1 score, or the Matthews correlation coefficient.

Finally, the integration of formal verification methods with data-driven approaches remains a significant gap, limiting the scalability and rigor of current validation frameworks for AV systems.

Many studies in AV research and simulations encounter common limitations (RQ7) that often mirror unresolved research gaps. These limitations frequently include narrow or unrepresentative datasets, computational constraints, difficulties in generalizing to real-world environments, and challenges in model complexity and interpretability. When studies fail to address these gaps, such as relying on overly simplified models or datasets that do not reflect real-world variability, they leave critical questions unexamined, diminishing the practical value of proposed solutions. For example, a model that performs well in idealized conditions but struggles in dynamic, real-world scenarios highlights a significant gap in its generalizability and robustness. To assess whether these gaps have been addressed, it is essential to consider how well the study adapts its methods to more complex, real-world contexts, including the incorporation of diverse datasets, real-time capabilities, and techniques aimed at improving model interpretability and scalability. If these challenges are not meaningfully tackled, the research remains constrained by its limitations, preventing it from making a substantial contribution to advancing AV technologies.

To address some of these gaps, the studies suggest a future research (RQ8) agenda. The field development requires a comprehensive approach to enhance the effectiveness of CC identification and validation. Standardizing definitions and taxonomies for CCs is crucial as it will allow for consistent comparisons across studies and improve the targeting of truly critical scenarios. Bridging the simulation-to-reality gap remains a central challenge. Advancements in sensor modeling, domain adaptation, and mixed-reality testing will increase the reliability of simulated CCs, making them more reflective of real-world conditions. One key step toward overcoming data scarcity is to create high-quality, community-shared datasets. Open-source initiatives such as the Udacity Self-Driving Car Dataset, Waymo Open Dataset, KITTI-360, and NuScenes already enable researchers to benchmark algorithms and train models on diverse, real-world data. However, despite the competitive environment where companies strive to prove their own autonomous models are superior, there is growing recognition that sharing safety-critical data and models can enhance overall safety. Similar to Volvo's proactive decision to open its intellectual property for the three-point seatbelt [192], which resulted in safer cars industry-wide, sharing AV data could have a profound societal impact. If such incentives are established through regulatory frameworks or collaborative industry initiatives, the pooling of safety-critical data could accelerate the development of robust testing frameworks and contribute to safer roads for everyone.

In conclusion, we hope that this SLR may have a significant practical relevance for industry stakeholders, regulatory bodies, and AV manufacturers. The adoption of automated CC identification tools could improve safety measures, while regulatory agencies may benefit from standardized approaches to evaluating AV systems. Manufacturers can apply these insights to refine their testing protocols and enhance real-world system reliability. Overall, while significant progress has been made in understanding and addressing CCs in AVs, continued efforts are essential to refine methodologies and ensure their effectiveness in real-world deployment.

## Acknowledgments

This work was carried out with the support of Itaú Unibanco S.A., through the Itaú Scholarships Program (PBI), linked to the Data Science Center at the Polytechnic School of the University of São Paulo.